\documentclass[lettersize,journal]{IEEEtran}
\usepackage{amsmath,amsfonts}
\usepackage{algorithmic}
\usepackage{algorithm}
\usepackage{array}
\usepackage{pdflscape}
\usepackage[caption=false,font=normalsize,labelfont=sf,textfont=sf]{subfig}
\usepackage{textcomp}
\usepackage{graphicx}
\usepackage{stfloats}
\usepackage{url}
\usepackage{verbatim}
\usepackage{graphicx}
\usepackage{booktabs}
\usepackage{cite}
\usepackage{tabularx}
\usepackage{tabulary}  
\usepackage{longtable}
\usepackage{tikz}
\usepackage{adjustbox}
\usepackage{tikz}
\usepackage{booktabs}
\usepackage{xcolor}
\usepackage{multirow}
\usepackage{academicons}
\usepackage{orcidlink}
\usepackage{pifont}
\newcommand{\cmark}{\ding{51}}

\colorlet{blue}{black}

\usetikzlibrary{shapes.geometric, positioning}
\begin{document}

\title{From Virtual Environments to Real-World Trials: Emerging Trends in Autonomous Driving}


\author{Aditya Humnabadkar*\,\orcidlink{0000-0002-9301-393X},~\IEEEmembership{Student Member,~IEEE},
        Arindam Sikdar*\,\orcidlink{0000-0002-5697-0060},~\IEEEmembership{Member,~IEEE},
        Benjamin Cave,
        Huaizhong Zhang\,\orcidlink{0009-0009-8899-4081},
        Nik Bessis\,\orcidlink{0000-0002-6013-3935},~\IEEEmembership{Senior Member,~IEEE},
        and Ardhendu Behera$^{\dag}$\,\orcidlink{0000-0003-0276-9000},~\IEEEmembership{Member,~IEEE}

\thanks{A. Humnabadkar, A. Sikdar, B. Cave, H. Zhang, N. Bessis and A. Behera authors are with the Department of Computer Science, Edge Hill University, United Kingdom.}
\thanks{*Equal contribution; $^{\dag}$Ardhendu Behera is the corresponding author, beheraa@edgehill.ac.uk}
\thanks{Manuscript received February xx, 2025; revised xx.}}

\markboth{Journal of \LaTeX\ Class Files,~Vol.~14, No.~8, August~2021}%
{Shell \MakeLowercase{\textit{et al.}}: A Sample Article Using IEEEtran.cls for IEEE Journals}


\maketitle

\begin{abstract}
Autonomous driving technologies have achieved significant advances in recent years, yet their real-world deployment remains constrained by data scarcity, safety requirements, and the need for generalization across diverse environments. In response, synthetic data and virtual environments have emerged as powerful enablers, offering scalable, controllable, and richly annotated scenarios for training and evaluation. This survey presents a comprehensive review of recent developments at the intersection of autonomous driving, simulation technologies, and synthetic datasets. We organize the landscape across three core dimensions: (i) the use of synthetic data for perception and planning, (ii) digital twin-based simulation for system validation, and (iii) domain adaptation strategies bridging synthetic and real-world data. We also highlight the role of vision-language models and simulation realism in enhancing scene understanding and generalization. A detailed taxonomy of datasets, tools, and simulation platforms is provided, alongside an analysis of trends in benchmark design. Finally, we discuss critical challenges and open research directions, including Sim2Real transfer, scalable safety validation, cooperative autonomy, and simulation-driven policy learning, that must be addressed to accelerate the path toward safe, generalizable, and globally deployable autonomous driving systems.
\end{abstract}


\section{Introduction}

\IEEEPARstart{A}{utonomous} Vehicles (AVs) have witnessed significant progress with deep learning methods enabling them to see, understand, and act within complex traffic environments. Modern perception systems can now interpret scenes with remarkable accuracy~\cite{blasch2021machine, thomas2020perception}. However, collecting large-scale real-world driving data is often costly, time-consuming, and constrained by safety and ethical considerations~\cite{muhammad2020deep}. Real-world datasets may lack diversity in rare or safety-critical scenarios, limiting an AV’s holistic testing. To address these gaps, researchers increasingly rely on simulation techniques to create virtual replicas of driving environments, often termed Digital Twins (DTs), which can be used to test and train AV models under a broader spectrum of conditions than would be feasible on real roads. During dataset creation, subtle differences in sensor noise, weather, and traffic behavior between virtual and physical domains create a domain gap~\cite{kar2019meta} that can degrade model performance when transferring from simulation to reality.

\begin{figure*}[htbp]
    \centering
    \includegraphics[width=0.7\textwidth]{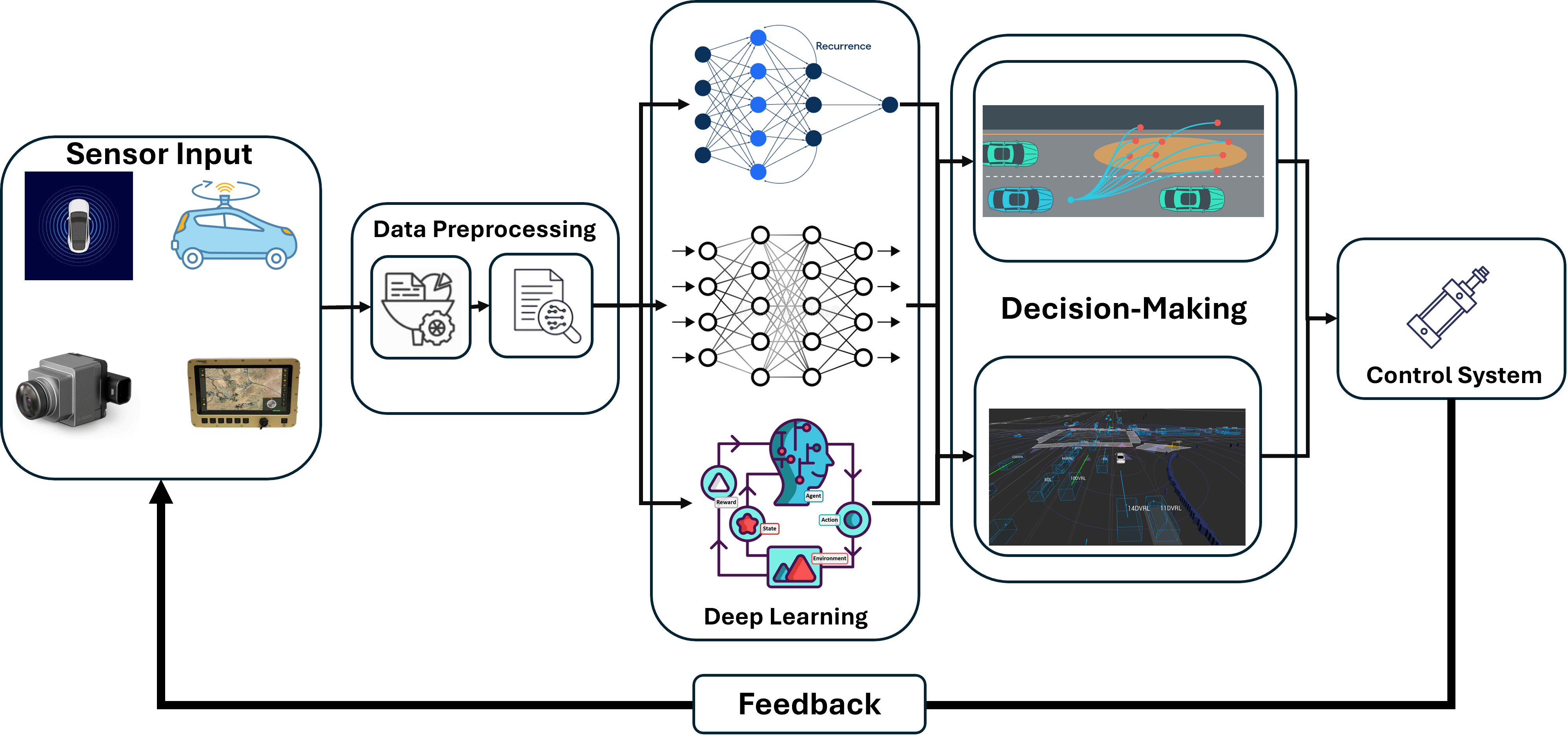}
    \caption{Illustration of a traditional AV perception-control pipeline. Sensor inputs (cameras, radar, LiDAR, GPS) capture environmental data, which is filtered and processed in a preprocessing stage. The refined data feeds into deep learning models for object detection and prediction. The outputs then guide high-level decision-making modules for path planning and behavior prediction, which are executed by the control system. A feedback loop continuously updates the system for real-time adjustments.}
    \label{fig:AV_system}
\end{figure*}

Within this landscape, Simulation-to-Reality (Sim2Real) and Reality-to-Simulation (Real2Sim) frameworks have become indispensable. These approaches transfer knowledge between simulated environments (where models are developed and tested) and the real world (where they are deployed) to iteratively reduce the domain gap. For example, Sim2Real techniques adapt models trained in simulators like CARLA~\cite{dosovitskiy2017carla}, Virtual KITTI~\cite{gaidon2016virtual}, or SYNTHIA~\cite{ros2016synthia} to perform reliably under unpredictable real-world conditions. Conversely, Real2Sim methods inject real-world data back into simulations, making virtual scenarios more faithful to reality~\cite{zhao2023real2sim2real}. By leveraging both directions, AV developers create feedback loops that continuously refine model performance (see Fig.~\ref{fig:AV_system}). In practice, a model trained under idealized virtual conditions might struggle with irregular lighting, surprise weather events, or unusual road user behavior when moved to real streets~\cite{zhao2023real2sim2real, sim2realcare}. Real2Sim alignment helps the simulator mirror such real-world challenges, ensuring that virtual testing uncovers issues before on-road deployment.

\begin{figure*}[htbp]
    \centering
    \includegraphics[width=0.7\textwidth]{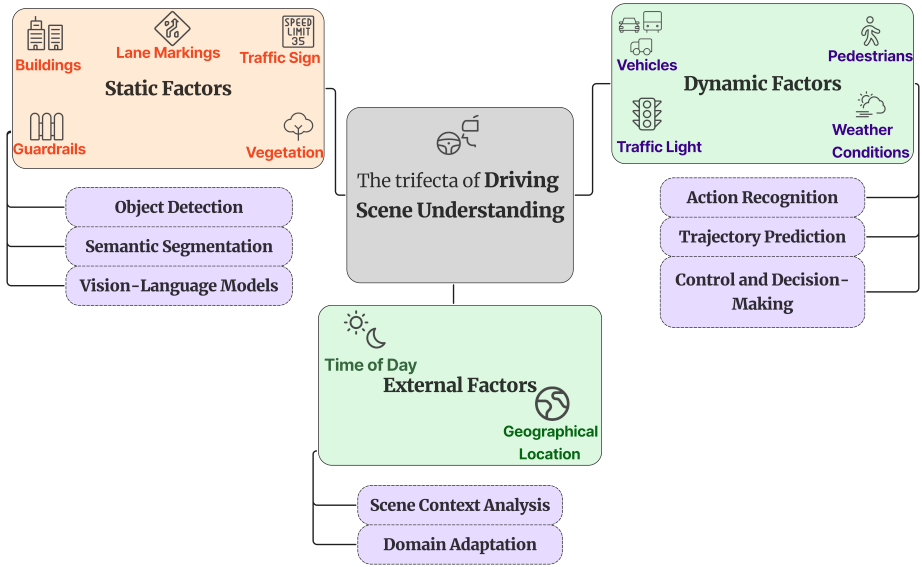}
    \caption{Illustration of the integration of static, dynamic, and external factors into a unified framework for driving scene understanding. Together, they inform perception tasks, guide decision-making processes, and enable domain adaptation strategies to handle the complexities of real-world environments. By simultaneously considering these three aspects, AVs can gain a comprehensive insight into their environment, leading to safer and more efficient navigation.}
    \label{fig:diagram3}
\end{figure*}

Another crucial aspect of understanding driving scenes is predicting the intentions and trajectories of other agents (vehicles, pedestrians, cyclists) for safe navigation. In virtual or mixed-reality settings, researchers can simulate a wide spectrum of human and vehicle maneuvers, enabling AI models to learn from both real and fabricated behaviors. This iterative cycle between real and synthetic data resources mitigates the domain gap and enhances model robustness~\cite{stocco2022mind}. Recently, Vision-Language Models (VLMs) have emerged as a promising tool to bridge simulated and real domains by embedding semantic information (textual cues) with visual data~\cite{zhang2024vision}. VLMs provide deeper contextual understanding of objects and scenes by associating images with descriptions or labels. For instance, real sensor imagery can be rendered in simulation and annotated with rich text, allowing models to align semantics across domains. By incorporating such multimodal understanding, self-driving systems become more resilient to domain shifts, gaining robust insights from diverse visual contexts whether physical or artificially generated.
\textcolor{blue}{Achieving robust scene understanding requires simultaneously modeling static infrastructure, dynamic agents, and external conditions through integrated perception architectures. Traditional CNNs established foundational capabilities: NVIDIA's PilotNet demonstrated end-to-end pixel-to-steering mapping~\cite{pilotnet}, while YOLO~\cite{yolo,yolov3} and SSD~\cite{ssd,aher2024advances,mozaffari2020deep,zhang2024deep,smith2023object} revolutionized real-time multi-class detection. PointPillars extended detection to 3D LiDAR point clouds~\cite{pointpillars}, complemented by Bird's-Eye-View (BEV) representations that unify multi-sensor fusion. Lift-Splat-Shoot encodes arbitrary camera configurations into BEV space~\cite{philion2020lift}, while BEVFusion fuses camera and LiDAR features for improved cross-domain robustness~\cite{liu2022bevfusion}.} Graph Neural Networks further enhance relational reasoning~\cite{luettin2022survey,commonroad,li2024survey,htun2024integrating}, with frameworks like CommonRoad-Geometric and PreGSU modeling inter-agent spatial-temporal interactions~\cite{commonroad,pregsu}. Semantic segmentation provides pixel-level scene parsing through efficient architectures (ENet, ICNet)~\cite{enet,icnet} and multi-scale models (DeepLabv3+, HRNet)~\cite{deeplab,hrnet}. Trajectory prediction employs RNNs and LSTMs with attention mechanisms~\cite{alahi2016,deo2018,zhang2021}, comprehensively surveyed for action recognition~\cite{grigorescu2020survey,chib2023recent,kong2022human}. Control strategies use deep reinforcement learning with attention-augmented architectures~\cite{kendall2019,lee2024instant,zhang2021,lee2022adas,lee2022stability,kiran2021deep,aradi2020survey,zhao2024survey}. Translating these capabilities to simulation demands careful handling of static (road layouts, signage), dynamic (vehicles, pedestrians), and external (weather, lighting) factors~\cite{rodus2024,uvda2022,d2nerf2022}, as illustrated in Fig.~\ref{fig:diagram3}. Neural-based 3D scene generation~\cite{wen20253d} enables procedural environment creation through PCG-based methods (CARLA~\cite{dosovitskiy2017carla}), neural-3D approaches (CityDreamer~\cite{xie2024citydreamer}, CityDreamer4D~\cite{xie2025compositional}), and video-based methods (MagicDrive~\cite{gao2024magicdrive}). Frameworks like DrivingGaussian employ Composite Gaussian Splatting for perspective-refined 3D representations~\cite{drivinggaussian}, while Vectorized Scene Representation enables real-time static feature editing~\cite{jiang2023vad}. Dynamic agent modeling combines LSTM/GNN architectures with VAE-GRU hybrids~\cite{deo2018,commonroad,drivinggaussian,lee2022stability,vaegru} and traffic management systems~\cite{li2021novel}. External factors are addressed through multimodal sensor fusion~\cite{huang2022multimodal}, comprehensive datasets (nuScenes)~\cite{qian2024nuscenes}, physics-based rendering, and domain randomization~\cite{zhao2024exploring}. This integrated treatment of perception architectures and simulation techniques is essential for creating high-fidelity virtual environments that accurately replicate real-world complexity.
In summary, developing reliable autonomous driving systems requires bridging the gap between real and simulated environments. This survey demonstrates how deep learning, vision language models (VLMs), and advanced simulation techniques collectively contribute to this objective, ultimately enabling more robust perception, prediction, and planning under diverse real world conditions. We also examine the ethical and logistical constraints associated with large scale real world data collection, which position synthetic data as a practical alternative for testing rare and safety critical scenarios. Simulation platforms provide controlled and repeatable testbeds to validate and refine autonomous vehicle algorithms across a wide range of traffic and weather conditions that would be prohibitively difficult or unsafe to replicate in physical trials~\cite{hu2022sim,revell2022sim2real}. Furthermore, we emphasize the importance of evaluating autonomous driving systems not only by conventional performance metrics but also against emerging safety standards and societal benchmarks such as ISO 21448 and OpenSCENARIO. Table~\ref{tab:review paper comparison} contrasts the scope of this survey with earlier works. While previous surveys have typically focused on individual themes such as synthetic datasets, simulation platforms, or domain adaptation, only a few have explored emerging topics like vision language models or digital twin frameworks in depth. Moreover, very few have connected these technical advancements to real world autonomous vehicle trials and deployment challenges. In contrast, this survey provides a unified and systematic overview that brings together simulation environments, multimodal learning, synthetic data generation, cooperative perception, and field testing. This broader perspective synthesizes existing knowledge while revealing new interconnections across research areas, addressing a critical gap in the current literature.

The remainder of the paper is organized as follows. Section~\ref{vision-language} discusses vision-language models for contextual scene understanding and how multimodal representations improve situational awareness and decision-making in AVs. Section~\ref{Section_taxonomy} presents a comprehensive taxonomy of driving scene datasets and critically examines the scarcity of rich multi-modal annotations, demonstrating how simulation-based approaches can address these fundamental data limitations. Section~\ref{sim-tech} examines simulation technologies that connect theoretical advancements to applied testing, including domain adaptation strategies. Section~\ref{sec:autonomus_driving} surveys real-world autonomous driving trials and their outcomes, and provides an overview of infrastructural, regulatory, and societal considerations for large-scale AV deployment. Finally, Section~\ref{discussion} discusses key insights, open research questions, and future directions toward an integrated, trustworthy autonomous driving ecosystem followed by conclusion in Section~\ref{conclusion}.
\begin{table*}[h!]
\centering
\caption{Comparison of Review Papers on Autonomous Driving Topics}
\renewcommand{\arraystretch}{1.2} 
\setlength{\tabcolsep}{6pt}      
\begin{tabular}{@{}p{7cm}cccccc@{}}
\toprule
\multirow{2}{*}{\textbf{Paper Title}} &
\multirow{2}{*}{\textbf{Synthetic Data}} &
\multirow{2}{*}{\textbf{Simulators}} &
\multirow{2}{*}{\textbf{VLMs}} &
\multirow{2}{*}{\shortstack{\textbf{Sim2Real/}\\\textbf{Real2Sim Transfer}}} &
\multirow{2}{*}{\textbf{Digital Twins}} &
\multirow{2}{*}{\textbf{AV Trials}} \\
& & & & & & \\
\midrule
\textit{How Simulation Helps Autonomous Driving: A Survey of Sim2real, Digital Twins, and Parallel Intelligence\cite{hu2023simulation}} 
& \cmark & \cmark & & \cmark & \cmark & \\
\textit{Synthetic Datasets for Autonomous Driving: A Survey\cite{song2023synthetic}} 
& \cmark & & & \cmark & & \\
\textit{Vision Language Models in Autonomous Driving: A Survey and Outlook\cite{zhou2024vision}} 
& & & \cmark & & & \\
\textit{A Survey on Autonomous Driving Datasets: Statistics, Annotation Quality, and a Future Outlook\cite{10509812}} 
& \cmark & & & & & \\
\textit{A Comprehensive Review on Traffic Datasets and Simulators for Autonomous Vehicles\cite{sarker2024comprehensive}} 
& \cmark & \cmark & & & & \\
\textit{Review and analysis of synthetic dataset generation methods and techniques for application in computer vision\cite{Paulin2023}} 
& \cmark & & & & & \\
\textit{A Survey on Datasets for the Decision Making of Autonomous Vehicles\cite{Paulin2023}} 
& \cmark & & & & & \\
\textit{Exploring the Sim2Real Gap Using Digital Twins\cite{sudhakar2023exploring}} 
& \cmark & & & \cmark & \cmark & \\
\textit{Digital Twins for Autonomous Driving: A Comprehensive Implementation and Demonstration\cite{wang2024digital}} 
& & \cmark & & & \cmark & \\
\textit{A Survey of Autonomous Driving: Common Practices and Emerging Technologies\cite{yurtsever2020survey}} 
& & & & & \cmark & \\
\textit{Explanations in Autonomous Driving: A Survey\cite{omeiza2021explanations}} 
& & & & & & \\
\textit{End-to-End Autonomous Driving: Challenges and Frontiers\cite{chen2024end}} 
& & & & \cmark & & \\
\textit{World Models for Autonomous Driving: An Initial Survey\cite{guan2024world}} 
& & \cmark & & & & \\
\textit{A Survey on Recent Advancements in Autonomous Driving Using Deep Learning\cite{zhao2024survey}} 
& \cmark & & & & & \\
\textit{Milestones in Autonomous Driving and Intelligent Vehicles: Survey of Surveys\cite{chen2022milestones}} 
& \cmark & & & & & \\
\textit{Perspective, Survey, and Trends: Public Driving Datasets and Toolkits for Autonomous Driving Virtual Test\cite{ji2021perspective}} 
& \cmark & \cmark & & & & \\
\midrule
\textbf{\textit{Our Survey}} 
& \cmark & \cmark & \cmark & \cmark & \cmark & \cmark \\
\bottomrule
\end{tabular}
\label{tab:review paper comparison}
\end{table*}
\begin{figure}[htbp]
    \centering
    \includegraphics[width=0.5\textwidth]{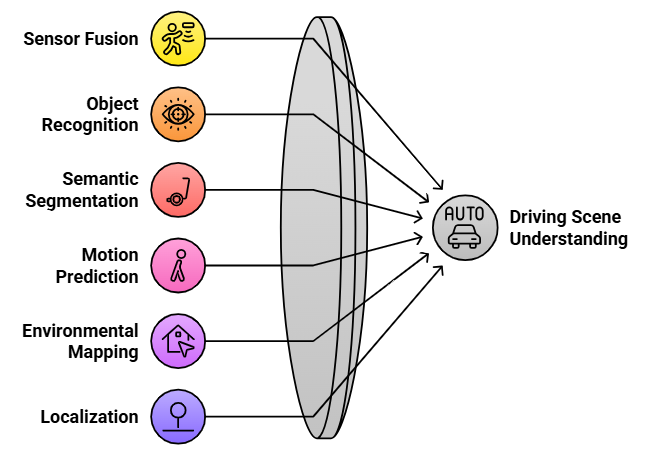}
    \caption{A unified framework for driving scene understanding where several critical components, namely, sensor fusion, object recognition, semantic segmentation, motion prediction, mapping, and localization – merge synergistically to enable robust and safe driving. Each component addresses specific perception, prediction, or decision-making challenges. Together, they allow the vehicle to navigate dynamic environments with precision and reliability.}
    \label{fig:scene_understanding}
\end{figure}

\begin{figure*}[htbp]
    \centering
    \includegraphics[width=1.0\textwidth]{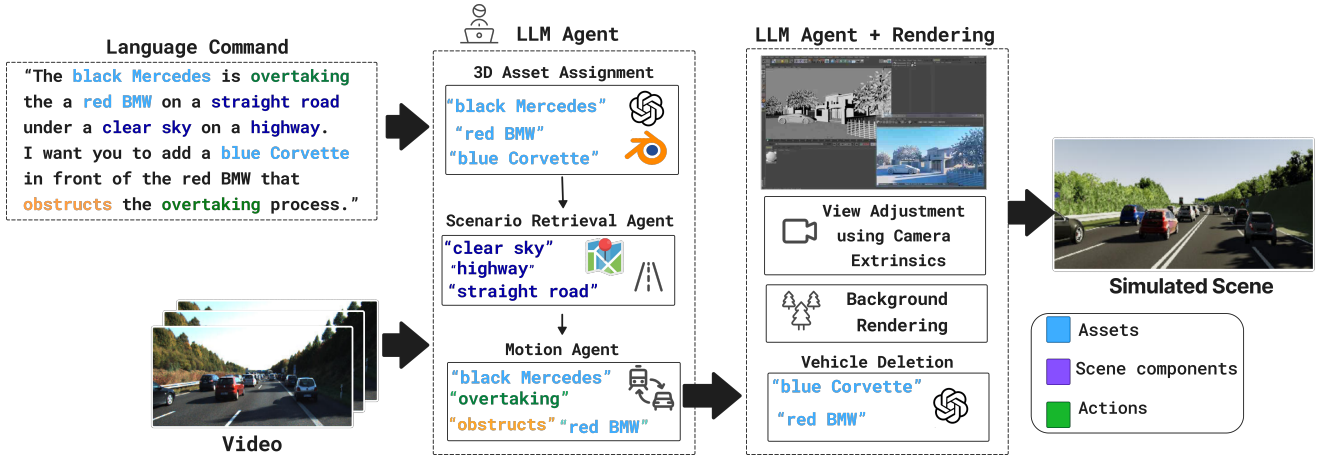}
    \caption{Showcasing a DriveGPT\cite{drivegpt4} workflow in which natural language inputs from users are interpreted by LLM agents to allocate 3D models, gather relevant contextual elements like the surroundings and movement dynamics, and execute rendering functions. It also includes view modification and background creation, ultimately producing a photorealistic driving scene with dynamic simulation. This workflow demonstrates the capability of integrating advanced language models with simulation tools to enable precise and flexible autonomous vehicle testing and scenario visualization, bridging the gap between human intent and machine-generated environments.}
    \label{fig:diagram5}
\end{figure*}
\section{Vision-Language Models (VLMs) for Contextual Scene Understanding}
\label{vision-language}
Vision-Language Models (VLMs) enable AVs to interpret both visual geometry and high-level semantics through integrated language reasoning. As illustrated in Fig.~\ref{fig:diagram5}, VLM-enabled systems can ground natural language commands ("overtake the red BMW") in visual perception for context-aware planning. Recent developments (Tables~\ref{table:VLMs},~\ref{table:VLMs_2}) span language-grounded navigation (Talk2Car~\cite{deruyttere2019talk2car}), LLM-driven reasoning (DiLu~\cite{wen2024dilu}, LanguageMPC~\cite{sha2023languagempc}), HD-map-free planning (NaVid~\cite{zhang2024navid}), camera-only systems (CarLLaVA~\cite{renz2024carllava}), 3D multimodal understanding (VLM2Scene~\cite{liao2024vlm2scene}, MAPLM~\cite{cao2024maplm}), zero-shot detection (CLIP2Scene~\cite{chen2023clip2scene}), and language-based trajectory generation (DriveGPT4~\cite{drivegpt4}). These approaches vary in environments, modalities, learning paradigms, and deployment readiness—early systems tackled constrained language grounding while recent models integrate large-scale reasoning and 3D spatial understanding.
\begin{table*}[htbp]
\centering
\caption{Evaluation of the different VLMs designed for scene understanding. 
State-of-the-art methods have been compared with respect to the datasets they use, 
the input modalities, learning types, target tasks, computational complexity, 
robustness, and generalization capabilities. All the models provide unique insights 
on using language as a medium to deepen the context of a scene. The only caveat is 
that there is no standardized application nor an appropriate metric that has been 
developed to accurately estimate the efficiency and usefulness of these models.}
\label{table:VLMs}
\resizebox{\textwidth}{!}{%
\begin{tabular}{|p{1.6cm}|p{1.3cm}|p{1.3cm}|p{1.3cm}|p{1.3cm}|
                p{1.6cm}|p{1.6cm}|p{1cm}|p{1cm}|p{1.3cm}|}
    \hline
    \textbf{Paper, Year} & \textbf{Env.} & \textbf{Input Modality} & \textbf{Output Modality}
      & \textbf{Learning Type} & \textbf{Technique} & \textbf{Key point} & \textbf{Versatility}
      & \textbf{Viability} & \textbf{Data} \\
    \hline
    Talk2Car (2019) \cite{deruyttere2019talk2car} 
      & Real-world & RGB camera, language instructions & Navigation adjustments 
      & Supervised learning & Visual-language grounding & Language-based interaction safety 
      & High & Moderate & Talk2Car dataset \\
    \hline
    DiLu (2024)\cite{wen2024dilu} 
      & Simulated, real-world & multimodal inputs (text, visual) & Decision-making commands 
      & Knowledge-driven reasoning & Explainability via reasoning modules 
      & Handles human-like reasoning in safety-critical tasks & Very High & High & Knowledge-based datasets \\
    \hline
    LanguageMPC (2023)\cite{sha2023languagempc} 
      & Various simulations & Language instructions, visual data & Control actions, navigation 
      & Language-to-control mapping & Attention maps from language to control 
      & Multi-vehicle coordination & High & Very High & Custom multi-agent datasets \\
    \hline
    LimSim++ (2024) \cite{fu2024limsim++} 
      & Urban simulations & RGB, language-based commands & Adaptive control actions 
      & Closed-loop multimodal learning & Insights through continuous adaptation 
      & Robustness across diverse scenarios & Moderate & Moderate & Multi-scenario simulated datasets \\
    \hline
    NaVid (2024)\cite{zhang2024navid} 
      & Real-world, simulation & Continuous RGB streams & Waypoints, navigational cues 
      & Vision-language navigation & Visual-textual mappings for navigation 
      & Navigates unseen environments & High & Moderate & Continuous RGB-based dataset \\
    \hline
    CarLLaVA (2024)\cite{renz2024carllava} 
      & CARLA, real-world & Camera (RGB), language input & Steering, throttle 
      & Vision-language integration & Auxiliary depth outputs, semantic maps 
      & High performance in safety-critical tasks & High & High & CARLA driving data \\
    \hline
    VLM2Scene (2024)\cite{liao2024vlm2scene} 
      & Synthetic datasets & Image, LiDAR, text & 3D scene understanding 
      & Self-supervised contrastive learning & Region-based semantic reasoning 
      & Detection in complex 3D environments & Very High & High & 3D synthetic datasets \\
    \hline
    NuScenes-QA (2024)\cite{qian2024nuscenes} 
      & NuScenes, Custom & RGB, multimodal inputs & Visual question answering 
      & multimodal supervised learning & Text-based scene reasoning 
      & Understands multimodal relationships & High & Moderate & NuScenes QA dataset \\
    \hline
    MAPLM (2024)\cite{cao2024maplm} 
      & Various environments & LiDAR, panoramic images, HD maps & multimodal scene understanding 
      & Cross-modal supervised learning & Cross-modal reasoning 
      & Handles complex cross-modal relationships & Very High & High & Large-scale traffic datasets \\
    \hline
    VLPD (2023)\cite{liu2023vlpd} 
      & KITTI, NuScenes & Multi-camera, LiDAR & Pedestrian detection 
      & Vision-language self-supervision & Activation maps for pedestrians 
      & Addresses occlusion and small object detection & High & Moderate & Pedestrian datasets \\
    \hline
    CLIP2Scene (2023)\cite{chen2023clip2scene} 
      & CARLA & RGB, depth, 3D point clouds & Zero-shot object detection 
      & Zero-shot learning & CLIP-based visual-language reasoning 
      & Robustness in data-sparse conditions & Very High & High & CARLA 3D datasets \\
    \hline
    DriveGPT4 (2023)\cite{drivegpt4} 
      & BDD-X & Language instructions, visual sensor data & Control predictions 
      & Instruction-tuned language modeling & Attention maps based on language input 
      & Handles complex navigation through reasoning & High & Very High & BDD-X dataset \\
    \hline
    VAD (2024)\cite{jiang2023vad} 
      & Synthetic environments & Vectorized scene data & Planning efficiency improvements 
      & Self-supervised vector learning & Instance-based representation analysis 
      & Efficient in computational resource use & Moderate & High & Vector datasets \\
    \hline
    CityLLaVA (2024)\cite{duan2024cityllava} 
      & Urban simulations & Bounding box, language prompts & Navigation adjustments 
      & Vision-language fine-tuning & Bounding box explanations 
      & Manages language-to-scene mapping challenges & High & Moderate & Urban simulation datasets \\
    \hline
    VLAAD (2024)\cite{park2024vlaad} 
      & BDD-X, HDD, DRAMA & RGB, LiDAR, natural language & Vehicle commands, navigation 
      & multimodal instruction tuning & Explainable through natural language 
      & Handles long-tail and rare driving conditions & Very High & High & VLAAD dataset \\
    \hline 
\end{tabular}
} 
\end{table*}
%
\begin{table*}[htbp]
\centering
\caption{(Continued) Additional VLMs for Scene Understanding}
\label{table:VLMs_2}

\resizebox{\textwidth}{!}{%
\begin{tabular}{|p{1.6cm}|p{1.3cm}|p{1.3cm}|p{1.3cm}|p{1.3cm}|
                p{1.6cm}|p{1.6cm}|p{1cm}|p{1cm}|p{1.3cm}|}
    \hline
    \textbf{Paper, Year} & \textbf{Env.} & \textbf{Input Modality} & \textbf{Output Modality}
      & \textbf{Learning Type} & \textbf{Technique} & \textbf{Key point} & \textbf{Versatility}
      & \textbf{Viability} & \textbf{Data} \\
    \hline

    DRAMA (2023)\cite{malla2023drama} 
      & Real-world & Camera, LiDAR, language & Risk localization 
      & Joint risk explanation & Text-based explanations for risks 
      & Identifies safety-critical objects & High & Moderate & Risk annotation datasets \\
    \hline
    VLP (2024)\cite{pan2024vlp} 
      & Simulated environments & Visual data, text & Planning tasks 
      & Vision-language planning & Text-embedded reasoning 
      & Supports challenging scenario planning & High & Moderate & Planning datasets \\
    \hline
    VLFM (2024)\cite{yokoyama2024vlfm} 
      & Custom scenarios & Vision-language mappings, spatial information & Zero-shot navigation 
      & Zero-shot navigation via language integration & Spatial-textual integration reasoning 
      & Effectiveness in unfamiliar environments & High & Moderate & Custom scenarios data \\
    \hline
    Natural Language Can Facilitate Sim2Real Transfer (2024)\cite{yu2024natural} 
      & Various environments & Language, RGB camera & Navigation adjustments 
      & Language-guided domain adaptation & Explainability through language-based cues 
      & Bridging Sim2Real gap & Very High & High & Domain adaptation datasets \\
    \hline
    DriveVLM-Dual (2023)\cite{tian2024drivevlm} 
      & Real-world, simulated & RGB, multimodal data & Trajectory planning 
      & multimodal integration & Explainability through trajectory reasoning 
      & Adaptability in dynamic driving scenarios & Very High & High & DriveVLM datasets \\
    \hline
    Vision-Language Frontier Maps (2024)\cite{yokoyama2024vlfm} 
      & Custom, real-world & Vision-language mappings, semantic context & Zero-shot object localization 
      & Semantic navigation via language & Spatial reasoning using language prompts 
      & Robustness in real-world navigation & Very High & High & Boston Dynamics Spot dataset \\
    \hline
    ChatScene (2024)\cite{zhang2024chatscene} 
      & Simulated, real-world & Language inputs, knowledge graphs & Safety-critical driving scenarios
      & Knowledge-based generation & LLM-driven scenario creation 
      & Combines domain knowledge and language models to create corner-case scenes 
      & High & Moderate & Synthetic + Real scenario datasets \\
    \hline
    Editable Scene Simulation (2024)\cite{wei2024editable} 
      & Simulated & Collaborative text instructions & Dynamically updated 3D scenes
      & Multi-agent learning & LLM-based scene editing 
      & Allows interactive editing of driving scenarios in simulation 
      & High & Moderate & Custom scene simulation dataset \\
    \hline
\end{tabular}
}
\end{table*}
\textcolor{blue}{Versatility assesses cross-domain applicability (environmental range, task generalizability); Viability evaluates deployment readiness (latency, hardware requirements). High viability indicates $\leq$200ms latency on consumer GPUs; Moderate requires offline processing or specialized hardware.}
Collectively, these approaches are compared across their environments (real or simulated), input and output modalities, learning types (supervised vs. self-supervised vs. knowledge-driven), and key innovations. We also note each model’s generalization ability, practical viability (e.g., inference speed), and the datasets used for development. A clear trend emerges as early VLM systems tackled direct language grounding in relatively constrained settings, whereas later systems integrated large-scale reasoning, 3D spatial understanding, and robust simulation ties. Critical analysis reveals deployment trade-offs: research models (DriveGPT4~\cite{drivegpt4}, DiLu~\cite{wen2024dilu}) achieve 2-5 FPS requiring \textgreater 16GB memory versus navigation-focused systems (VLFM~\cite{yokoyama2024vlfm}, NaVid~\cite{zhang2024navid}) reaching 15-25 FPS with reduced reasoning depth. CarLLaVA~\cite{renz2024carllava} achieved 32.6\% performance gains but requires RTX 4090 GPUs with 500ms-2s delays, exceeding safe driving's ~100-200ms budget. Robustness remains challenging: 20-30\% performance drops in low-light, 15-25\% reductions in heavy rain. While conceptually powerful, practical deployment requires efficient architectures, hardware acceleration, and domain-specific optimizations to balance real-time constraints with reasoning capabilities. Success would enable interactive scenario testing where engineers query AVs in natural language, bridging human insight with machine precision for safer systems.

\textcolor{blue}{World models show promise for future state prediction and synthetic video generation, with approaches like Cosmos-Transfer1~\cite{nvidia2025cosmos} enabling conditional multimodal control. A recent survey~\cite{gao2025foundationmodels} comprehensively covers generative AI for sim2real tasks. However, VLM application to controllable corner-case generation and causal reasoning requires further exploration.} The next section examines how simulation addresses VLM limitations through infinite data variation and edge-case testing.
\begin{figure*}[htbp]
    \centering
    \includegraphics[width=0.87\linewidth]{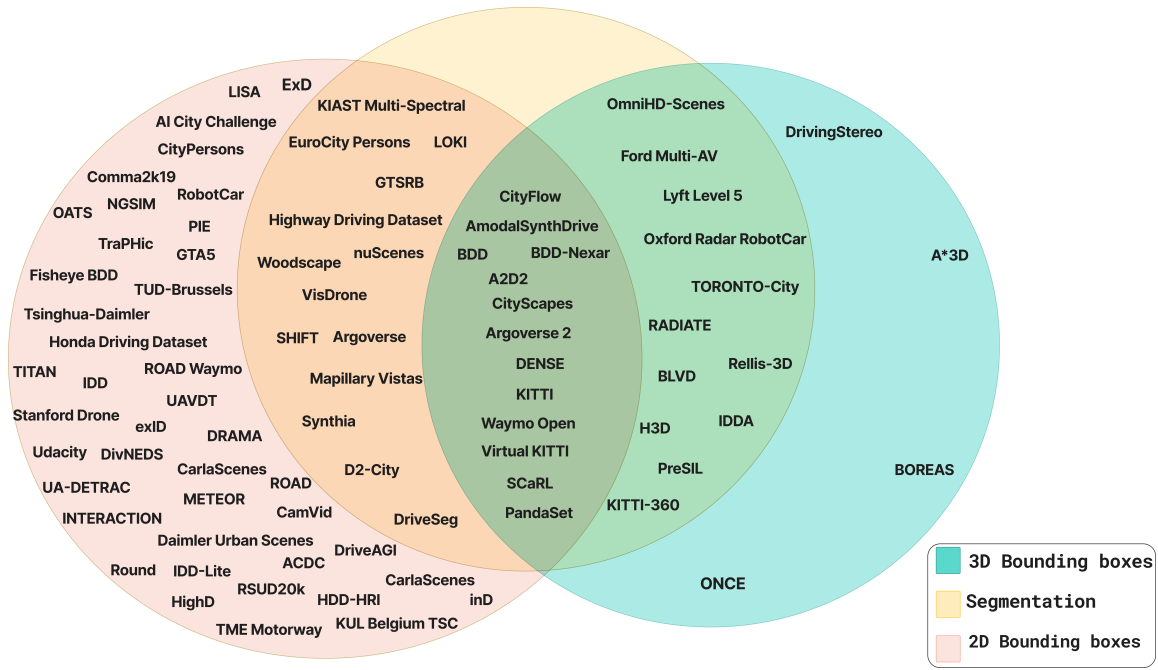}
    \caption{The Venn diagram shows the distribution of recent driving datasets across annotation types  . About 50–60\% of datasets provide only 2D bounding boxes; 20–30\% include 3D bounding boxes along with other annotations; nearly 20\% include segmentation masks (again typically alongside other modalities). Only a small fraction (5–10\%) of datasets contain all three annotation types. Moreover, accounting for static, dynamic, and external factors makes data collection and labeling even more challenging. This underscores the need to supplement real datasets with simulation-based pipelines to cover rare or complex scenarios.}
    \label{fig:diagram6}
\end{figure*}
\section{Annotation-Based Taxonomy of Autonomous Driving Datasets}
\label{Section_taxonomy}
Robust autonomous driving systems critically depend on comprehensive datasets that effectively represent real-world complexity. An essential step in bridging the domain gap between simulation and reality involves meticulously categorizing driving datasets according to their annotation types. Such categorization allows researchers and practitioners to identify gaps, understand dataset capabilities, and select appropriate resources for developing advanced perception models. Fig.~\ref{fig:diagram6} illustrates a taxonomy of recent driving datasets categorized by their annotation types, specifically, 2D bounding boxes, segmentation masks, and 3D bounding boxes and highlights significant coverage gaps in dataset annotations. While comprehensive perception tasks ideally require datasets annotated simultaneously across these three modalities, fewer than 10\% of datasets surveyed offer such completeness.

\paragraph{Datasets with 2D Bounding Box Annotations}
The most prevalent dataset category features 2D bounding box annotations, reflecting the fundamental object detection task in autonomous driving. These annotations typically comprise rectangular bounding boxes around objects such as vehicles, pedestrians, and cyclists within images, providing coarse spatial and class information. Datasets like KITTI~\cite{geiger2012we} have historically established benchmarks for vision-based object detection, significantly influencing perception algorithm development. Similarly, large-scale datasets such as BDD100K~\cite{yu2020bdd100k} provide diverse on-road video frames annotated with 2D bounding boxes across various object classes, geographical regions, and environmental conditions. Although 2D bounding boxes are relatively simple to annotate, their informational depth remains limited, capturing only object presence and approximate location without precise shape or spatial context.

\paragraph{Datasets with Segmentation Annotations}
Datasets offering segmentation annotations provide granular, pixel-level scene understanding. Semantic segmentation delineates each pixel into distinct classes (e.g., roads, vehicles, pedestrians), thus supporting advanced perception tasks including scene parsing and environmental mapping. Cityscapes~\cite{cordts2016cityscapes} represents a foundational semantic segmentation dataset, featuring pixel-level annotations of complex urban environments, while Mapillary Vistas provides global coverage with a diverse array of object classes. Segmentation annotations significantly enrich scene understanding but entail substantial annotation efforts, explaining their relative scarcity and smaller scales compared to bounding-box-only datasets.

\paragraph{Datasets with 3D Bounding Box Annotations}
Datasets annotated with 3D bounding boxes support comprehensive spatial reasoning by providing precise three-dimensional positions, sizes, and orientations of objects. Typically aligned with multi-sensor setups involving LiDAR and stereo cameras, such datasets like nuScenes~\cite{caesar2020nuscenes}, Waymo Open Dataset~\cite{sun2020scalability}, Argoverse~\cite{chang2019argoverse}, and Lyft Level 5 facilitate robust multi-sensor fusion and advanced 3D detection algorithms. These datasets typically accompany synchronized 2D annotations but rarely include segmentation labels due to prohibitive annotation costs because manual 3D labeling requires skilled labor and specialized tools~\cite{kappel2024pretraining, chen2023rebound}. Furthermore, capturing safety-critical yet infrequent scenarios (emergency vehicle interactions, extreme weather) remains challenging in real-world collection, motivating simulation-based augmentation approaches discussed in Section~\ref{sim-tech}.
\begin{table*}[htbp]
\centering
\caption{Technical Comparison of Modern Autonomous-Driving Simulators}
\label{tab:modern-simulators}
\resizebox{\textwidth}{!}{%
\begin{tabular}{
|p{2.2cm}  
|p{2.0cm}  
|p{2.5cm}  
|p{2.0cm}  
|p{2.5cm}  
|p{1.8cm}  
|p{2.5cm}  
|p{1.0cm}  
|p{2.0cm}  
|p{2.5cm}  
|}
\hline
\textbf{Simulator} & \textbf{Physics / Graphics} & \textbf{Sensor Fidelity / Realism} & \textbf{Scenario \& Traffic} & \textbf{Weather \& Env.\ Variation} & \textbf{ROS Integration} & \textbf{Multi-Agent \& Parallelism} & \textbf{Open Source} & \textbf{Certification / Compliance} & \textbf{Key Technical Highlights} \\
\hline
\textbf{CARLA}~\cite{dosovitskiy2017carla} &
Unreal Engine, custom physics pipeline &
Camera (RGB, depth), LiDAR, Radar, GPS, IMU &
Built-in Traffic Manager; Scenario Runner &
Dynamic weather, time-of-day changes, varying road conditions (wet/dry, fog) &
Native ROS bridge (ROS1, partial ROS2) &
Multiple vehicles/agents in synchronous or real-time mode; large-scale HPC possible with manual setup &
Yes &
N/A (Research) &
High-fidelity urban maps, realistic traffic logic, flexible Python APIs, large open-source community \\
\hline
\textbf{SVL Simulator} (\emph{formerly LGSVL})~\cite{rong2020lgsvl} &
Unity with modular physics plug-ins &
Camera, LiDAR, Radar, GPS, IMU &
Scenario Editor &
Dynamic weather and lighting changes &
Native support (ROS1 and ROS2) &
Scenario Editor &
Yes &
N/A (Research) &
Integrated Autoware and Apollo AD stacks; user-friendly scenario creation and sensor configuration; active community \\
\hline
\textbf{NVIDIA DRIVE Sim}~\cite{nvidia_drivesim} &
Omniverse and PhysX &
Photorealistic camera feeds (with sensor artifacts), LiDAR, Radar &
Scenario authoring via Omniverse &
Physically based sky system, dynamic lighting, advanced reflections and refractions for weather variability &
Closed-source but adapters exist for ROS/ROS2 integration &
Highly parallelizable on GPU clusters; enterprise HPC pipelines for large-scale synthetic data generation &
Partial &
Enterprise (not certified) &
Real-time photorealism powered by GPU ray tracing, robust sensor simulation, strong integration with NVIDIA's AD software stack \\
\hline
\textbf{rFpro}~\cite{rfpro_software} &
Proprietary motorsport-grade engine &
High-fidelity camera feeds, LiDAR, Radar &
AI traffic vehicles and real-time scenario definition, SUMO integration &
Physically accurate weather and surface conditions (wet track, ice, mud), day/night transitions &
Limited (via proprietary bridges) &
Optimized for real-time HPC and Hardware-in-the-Loop (HIL); multi-vehicle concurrency common in OEM testing &
No &
OEM validation (not ISO 26262) &
Ultra-accurate vehicle dynamics (e.g.\ tire modeling), widely used by automotive OEMs; ideal for track-level precision and HIL \\
\hline
\textbf{dSPACE}~\cite{pomerantz2009dspace} &
ISO 26262-certified simulation platform &
High-fidelity sensor models &
Scenario-based testing &
Configurable weather models &
Native support &
Hardware-in-the-Loop (HiL/SiL) &
No &
ISO 26262 certified &
Industry standard for functional safety validation, OEM-grade HiL workflows \\
\hline
\textbf{aiSim}~\cite{aisim2024} &
Certified simulation framework &
Camera, LiDAR, Radar &
Traffic scenario generation &
Weather variation support &
Integration available &
Scalable multi-agent &
No &
ISO 26262 certified &
Used for ADAS validation, regulatory compliance testing \\
\hline
\textbf{MORAI}~\cite{morai_simulator} &
Proprietary graphics \& physics &
Camera, LiDAR, Radar, GPS, IMU &
Advanced traffic behaviors in digital twin city maps; multi-agent scenario setup &
Day/night cycles, rain, fog &
Limited &
Scalable to full city environments &
No &
N/A &
Emphasis on digital twin fidelity, realistic sensor/dynamic models, enterprise-level support for large-scale city simulations \\
\hline
\textbf{BeamNG.tech}~\cite{beamng_tech} &
BeamNG's soft-body physics engine &
Primarily camera-based simulation; LiDAR possible via third-party plug-ins &
Proprietary AI controlled traffic &
Limited weather changes (sunny, rain, fog) with basic environment variation &
Partial (via APIs or bridging) &
Multiple vehicles run in real time, but HPC-scale parallelism requires manual setup &
No &
N/A &
Unparalleled collision realism and soft-body dynamics, widely used for crash scenarios and advanced vehicle kinematics research \\
\hline
\textbf{Scenario Gym}~\cite{scenariogym2024} & 
Lightweight engine & 
Basic camera, configurable sensors & 
Scenario-centric execution & 
Limited & 
ROS compatible & 
Single-agent focus & 
Yes &
N/A (Research) &
Minimal setup, fast iteration, scenario-focused testing \\
\hline
\textbf{GarchingSim}~\cite{garchingsim2024} & 
Unity-based with photorealistic rendering & 
Camera, LiDAR & 
Simplified scenario definition & 
Basic weather variation & 
ROS integration & 
Limited multi-agent & 
Yes &
N/A (Research) &
Photorealistic output, minimalist workflow, academic-friendly \\
\hline
\end{tabular}%
}
\end{table*}
\begin{figure*}[h]
    \centering
    \includegraphics[width=1.0\textwidth]{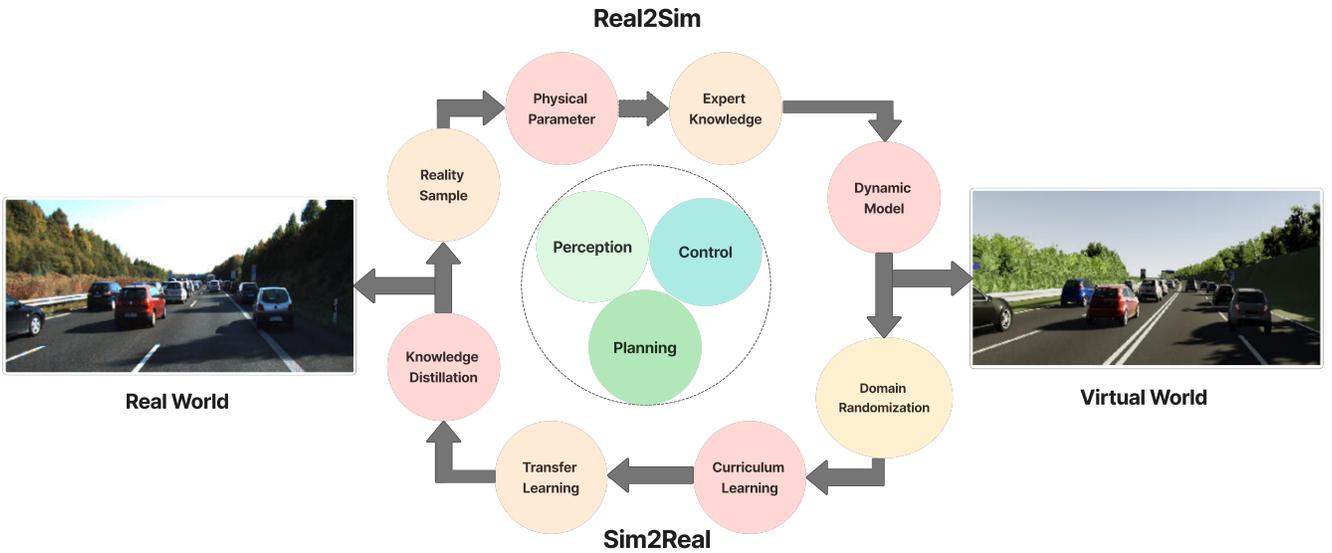}
    \caption{A cyclical pipeline of Real2Sim and Sim2Real bridges the domain gap between synthetic training environments and real‐world conditions. In the phase of Real2Sim, the physical parameters and expert knowledge have been encoded into dynamic models that are enriched by domain randomization. Similarly, in the Sim2Real, the adaptation of the models back to real‐world sensor inputs is done using curriculum learning, transfer learning, and knowledge distillation. The iteration of this loop progressively reduces the gap, enhancing robustness and operational safety in autonomous driving.}
    \label{fig:sim2real}
\end{figure*}
\section{Simulation Technologies: Bridging Theory and Practice} \label{sim-tech}
Simulation technologies are essential for developing and validating autonomous vehicle (AV) systems, offering controlled, repeatable, and safe environments that bridge the gap between theoretical models and real-world deployment. This section explores key simulation methodologies, including Sim2Real and Real2Sim transfer, digital twin-based testing, and comparative analyses of simulation platforms and domain adaptation strategies. Collectively, these approaches highlight how virtual testing accelerates AV development while supporting real-world applicability.

\subsection{The Role of Sim2Real and Real2Sim in Generating Complex Driving Scenarios}
Sim2Real and Real2Sim are core paradigms for transferring knowledge between simulation and reality in autonomous driving. Sim2Real involves training models in virtual environments and deploying them in the real world. Platforms such as \textit{CARLA}\cite{dosovitskiy2017carla}, \textit{Virtual KITTI}\cite{gaidon2016virtual}, and \textit{SYNTHIA}\cite{ros2016synthia} simulate urban driving scenes with diverse conditions, enabling scalable training (see Table~\ref{tab:modern-simulators}). However, models trained in simulation often underperform in reality due to the domain gap caused by visual and behavioral differences\cite{sim2realcare, zhao2023real2sim2real, stocco2022mind}. These gaps result from unmodeled real-world complexity, such as unpredictable lighting, rare events, or non-compliant road users. Real2Sim takes the reverse approach, using real-world data to refine simulation environments. This includes importing sensor noise characteristics, traffic patterns, and trajectory logs to improve realism and support scenario replay~\cite{zhao2023real2sim2real}. Together, Sim2Real and Real2Sim enable a complementary feedback loop, where simulation aids scalable training, and real data calibrates and validates the virtual environment.

\subsubsection{Addressing the Domain Gap Challenge}
A key challenge in transferring models from simulation to the real world is the domain gap, defined as the difference in data distributions between the two domains. This gap typically arises from two sources: (a) the appearance gap, involving visual discrepancies such as missing textures, lighting effects, and sensor artifacts~\cite{sim2realcare}, and (b) the content gap, which refers to differences in scene layout, object interactions, and behavioral patterns. Simulated environments often lack the unpredictability of real-world traffic, including jaywalking pedestrians or irregular driving behaviors, which are difficult to reproduce in rule-based simulations~\cite{stocco2022mind, zhao2023real2sim2real}. Formally, this gap can be expressed as the difference between the joint distributions of sensor inputs $X$ and outputs $Y$ in simulation and real domains:
\[
\Delta P = P_{sim}(X, Y) - P_{real}(X, Y)
\]
A larger $\Delta P$ indicates greater performance degradation when transitioning across domains. For instance, a model trained solely on clear, daytime simulations may fail under dusk lighting (appearance gap) or struggle to detect a child running across the street if such behavior was absent in training data (content gap). To mitigate this gap, several strategies have been proposed. Feature alignment methods, such as CARE~\cite{sim2realcare}, align simulated and real data distributions using adversarial training or statistical matching, often with class-specific reweighting. High-fidelity digital twins offer geographic realism by replicating real-world infrastructure and traffic flow~\cite{zhao2023real2sim2real}. In control systems, Sim2Real for MPC employs system identification and robust control to address discrepancies in vehicle dynamics and surface friction~\cite{stocco2022mind}. \textcolor{blue}{Domain randomization introduces variation in visual and structural elements to promote invariance. Quantitative performance comparisons of these strategies, including domain randomization (8-12\% gains), CARE (18\% improvement), and supervised fine-tuning (90\% gap closure). Recent work has critically examined evaluation metrics for domain transfer. Lambertenghi et al.~\cite{lambertenghi2024assessing} demonstrated that commonly used metrics such as FID scores are not reliable indicators of neural reality gap mitigation, proposing alternative quality metrics that better correlate with downstream task performance in autonomous driving testing.}

\subsubsection{Generating Robustness in Real2Sim and Sim2Real}
\textcolor{blue}{Robustness in domain transfer depends on both simulation augmentation and iterative integration of real-world feedback. Diffusion models have recently emerged as powerful tools for domain augmentation; Baresi et al.\cite{baresi2025efficient} demonstrated efficient test set augmentation for autonomous driving using diffusion-based image synthesis, enabling targeted generation of challenging scenarios for model validation. In Real2Sim, systems like AutoDRIVE\cite{samak2023towards} embed real-time data streams into simulations, enabling continuous model validation under realistic conditions. A more advanced framework, Real2Sim2Real, iteratively refines both simulation fidelity and model robustness. Techniques such as neural surface reconstruction improve virtual asset realism based on real-world trials, which are then used to retrain control policies in simulation before re-deployment. Over successive iterations, the simulation environment incorporates rare and safety-critical patterns, increasing coverage and training efficiency. Studies indicate that beyond a certain volume of real data (e.g., 10,000 labeled frames), the marginal utility of further annotation diminishes~\cite{salvato2022sim}. Simulation then becomes the primary driver of data diversity, enabling robust AV models through a balanced integration of real-world grounding and synthetic scalability.}

\subsection{Digital Twins: A Comprehensive Approach to AV Testing}
Digital Twins (DTs) are high-fidelity virtual replicas of physical systems that remain synchronized with their real-world counterparts in real time. In the context of autonomous driving, a DT mirrors critical aspects of both the vehicle and its environment, enabling developers to evaluate hypothetical scenarios in simulation while the physical vehicle operates normally or remains stationary. This approach facilitates safe and scalable testing, allowing detailed analysis of system responses without physical risk. DTs model complex interactions between hardware components such as sensors and actuators and software modules responsible for perception, planning, and control. They are especially valuable for assessing system behavior in rare or hazardous scenarios, including sudden pedestrian crossings or low-visibility conditions. Real-time data integration is essential for effective DT performance. Projects such as Tokyo Tech’s smart mobility field~\cite{wang2024smart, digitaltwinsaed2023} utilize edge computing and cloud infrastructure to stream live sensor data, including LiDAR and camera inputs, into the simulation. This ensures the twin maintains close alignment with real-world conditions, supporting synchronized perception and planning evaluations. Feedback from the DT can guide offline development or inform the AV’s real-time decision-making process. Table~\ref{tab:DT_Comparison} compares various DT implementations, showing trade-offs across scenario complexity, synchronization accuracy, processing latency, and computational requirements. While full-scale DTs offer rich simulation capabilities, they often demand significant infrastructure, whereas lighter DTs focus on specific monitoring tasks and require fewer resources. Beyond their role in real-time mirroring, DTs support a wide range of autonomous vehicle development workflows. In particular, they enable structured scenario-based testing and contribute significantly to bridging the gap between simulated and real-world data distributions. These two complementary capabilities are discussed below.

\subsubsection{Scenario-Based Testing with Digital Twins}
Scenario-based testing is a key application of DTs, providing a structured alternative to broad or randomized testing. Developers can define specific high-risk scenarios, such as highway merging or emergency braking, and run them repeatedly in simulation. Frameworks like PanoSim~\cite{digitaltwinsaed2023} support this by combining real vehicle data with virtual scenarios. For example, a twin vehicle can be placed in an emergency braking situation while the physical vehicle drives under normal conditions. This enables exhaustive testing of rare or unsafe situations, allowing developers to adjust perception or control parameters and re-evaluate performance without physical risk. Hazardous conditions, including near-collisions, heavy rainfall, or low traction, can also be safely explored in the twin, helping to identify and address potential system weaknesses.

\subsubsection{Digital Twins and Domain Gap Mitigation}
In addition to structured testing, DTs play a significant role in narrowing the gap between simulation and reality. Because DTs are continuously updated with real-world data, they serve as dynamic and accurate models of the current environment. This form of Real2Sim adaptation helps ensure that simulation scenarios reflect live operating conditions. For instance, a city-scale DT may integrate traffic camera feeds, weather reports, and IoT data to maintain an up-to-date view of congestion, road closures, and environmental hazards. When AV algorithms are validated in such an environment, the results are more likely to translate effectively to real-world performance. Studies have demonstrated the benefits of DTs that integrate LiDAR localization, roadside unit inputs, and cloud processing to support path planning and dynamic rerouting~\cite{digitaltwinconcepts2022}. One deployment used live updates to run AV decision-making algorithms inside the twin, continuously evaluating behavior under real-time conditions~\cite{tokyoedgecomputing}. This strategy enables early detection of system flaws and supports rapid model refinement. \textcolor{blue}{As DT technology advances, its utility for validation, domain adaptation, and safe deployment will continue to grow.}

\textcolor{blue}{\subsubsection{Neural Reconstruction for AV Validation}
Neural rendering and reconstruction techniques are increasingly being applied to autonomous driving simulation, offering photorealistic sensor data synthesis from real-world captures. RealEngine~\cite{jiang2025realengine} demonstrates simulating autonomous driving within reconstructed realistic contexts, enabling closed-loop testing in digitized real environments. Hybrid approaches such as that proposed by T\'{o}th et al.~\cite{toth2025hybrid} merge neural and physics-based rendering for multimodal sensor simulation, combining the visual fidelity of neural methods with the physical accuracy of traditional simulation.}

\textcolor{blue}{Industry adoption of neural reconstruction is accelerating rapidly. NVIDIA's Neural Reconstruction Engine (NuRec) leverages neural radiance fields for AV simulation, while aiSim World Extractor enables automatic scene reconstruction from recorded sensor data. Applied Intuition's Neural Sim and Waabi's simulation platform similarly employ neural reconstruction to achieve unprecedented realism in synthetic sensor generation. These tools represent a paradigm shift from handcrafted virtual environments toward data-driven scene reconstruction, significantly reducing the domain gap between simulation and reality.}

\begin{table*}[htbp]
\centering
\caption{Comprehensive Technical Comparison of Digital Twin Implementations in Autonomous Driving}
\renewcommand{\arraystretch}{1.3} 
\resizebox{\textwidth}{!}{%
\begin{tabular}{|p{2.3cm}|p{2.3cm}|p{3.5cm}|p{2.5cm}|p{2.3cm}|p{2.5cm}|p{3cm}|}
\hline
\textbf{Study} & \multicolumn{2}{|c|}{\textbf{Digital Twin}} & \textbf{Scenario Complexity} & \multicolumn{2}{|c|}{\textbf{Performance}} & \textbf{Applications} \\
\cline{2-3} \cline{5-6}
& \textbf{Type} & \textbf{Sync Capability} & & \textbf{Real-Time Processing} & \textbf{Computational Req.} & \\
\hline
Duan et al.~\cite{digitaltwinsaed2023 } & PanoSim-Based Twin & Integration of real vehicles and virtual environments & High: End-to-end testing for AV systems & High & Moderate: Requires edge computing & Autonomous vehicle testing and evaluation \\
\hline
Samak et al.~\cite{samak2023autodrive} & Open Ecosystem Twin & Synchronization of simulated and real-world data & Medium: Autonomous parking and driving tasks & Moderate & Moderate: Lightweight models & Development and deployment of autonomous driving algorithms \\
\hline
Zhang et al.~\cite{shoukat2022application} & Test Platform Digital Twin & Combined use of simulation and real-world testing tools & High: Scenario-specific testing environments & High & Moderate: Optimized for validation setups & Testing and validation of intelligent vehicles \\
\hline
Liang et al.~\cite{li2024platform} & Platform-Agnostic Digital Twin & Deep RL framework supporting various platforms & High: Lane keeping, multi-lane decision-making & High & High: GPU-accelerated training & Reinforcement learning-based decision-making \\
\hline
Klar et al.~\cite{klar2024maturity} & Monitoring and Control Twin & Data-driven vehicle condition monitoring & Medium: Predictive analysis for AV control & Medium & Moderate: Minimal hardware requirements & Vehicle monitoring, fault prediction, and autonomous driving \\
\hline
Voogd et al.~\cite{voogd2023reinforcement} & Vehicle-Centric Twin & Bi-directional synchronization of vehicle dynamics & High: Path following and steering control & High & High: Integration of real and simulated data & Reinforcement learning for vehicle dynamics \\
\hline
Wang et al.~\cite{digitaltwinsaed2023} & Infrastructure-Centric Twin & Real-time traffic updates using RSUs and edge computing & Medium: City-level traffic and road networks & High & Moderate: Edge computing and cloud processing & Traffic flow optimization and route planning \\
\hline
Sudhakar et al.~\cite{sudhakar2023exploring} & Adversarial Digital Twin & Robust domain transfer for adverse conditions & High: Adverse weather scenarios & High & High: Cloud-based simulation & Adversarial scenarios for stress testing \\
\hline
He et. al.~\cite{he2024advanced} & Generative Scene Twin & Language-driven 3D scene and scenario generation & High: User-defined dynamic scenarios & High & Very High: GPU-intensive rendering & Scenario generation, visualization, AV testing \\
\hline
\end{tabular}%
}
\label{tab:DT_Comparison}
\end{table*}

\subsection{Critical Comparison and Trade-off Analysis} \label{sec:trade-off}
While earlier sections detail the capabilities of simulation platforms and domain adaptation techniques, effective deployment of autonomous vehicle (AV) systems requires a deeper understanding of their performance trade-offs. This section critically examines leading simulation tools and domain adaptation strategies, emphasizing practical limitations, comparative strengths, and guidance for selecting technologies across the AV development pipeline.
\subsubsection{Simulation Platform Performance Trade-offs}
\textcolor{blue}{Simulation platforms differ in fidelity, scalability, cost, and applicability. Open-source tools like CARLA~\cite{dosovitskiy2017carla} are popular in academia due to their flexibility and low cost. CARLA supports high-fidelity urban scenarios but experiences performance bottlenecks under load (e.g., frame rates fall with $\sim$30 vehicles), and domain gaps remain, with real-world transfer performance degrading by up to 20\%. By contrast, NVIDIA DRIVE Sim~\cite{nvidia_drivesim} delivers photorealistic rendering with sensor modeling errors as low as $\sim$2 pixels, reducing the Sim2Real gap to 5–10\%. However, its reliance on high-end GPUs increases infrastructure cost and energy consumption. At the high end, rFpro~\cite{rfpro_software} is used in certification workflows, offering $<$3\% dynamics error and hardware-in-the-loop compatibility. It excels in controller validation but is less suited for perception model development and requires significant capital investment. Each simulator serves a specific phase: CARLA for rapid prototyping, DRIVE Sim for realistic validation, and rFpro for regulatory testing. This phased use optimizes cost, realism, and technical suitability throughout development.}
\subsubsection{Domain Adaptation Strategy Analysis}
Bridging the sim–real domain gap involves techniques such as domain randomization, feature alignment, and fine-tuning. Domain randomization improves robustness by exposing models to diverse simulated conditions, offering 8–12\% accuracy gains. However, excessive variation can degrade performance due to unrealistic training data. Feature alignment techniques like CARE~\cite{sim2realcare} offer more targeted improvements, yielding up to 18\% better transfer accuracy while reducing training overhead. Yet, they require careful tuning of adaptation losses and sometimes labeled real data. Semi-supervised fine-tuning with a small real dataset often closes 90\% of the domain gap, compared to 70–80\% for unsupervised methods, though it increases computational cost significantly. Hence, layered adaptation, consisting of randomization followed by alignment and light fine-tuning, emerges as an effective strategy.
\subsubsection{Deployment Readiness Assessment}
Deployment readiness varies across platforms and applications. CARLA dominates academic use, while NVIDIA DRIVE Sim and rFpro are more common in industry and regulatory contexts. However, several bottlenecks remain. Integrating vision-language models introduces latency (500 ms–2 s), limiting real-time applicability. Sensor simulation is also uneven: camera models are mature, but radar and LiDAR simulations remain less accurate under adverse conditions. Economic constraints further influence readiness. High-fidelity simulation infrastructure can cost millions annually~\cite{nvidia_drivesim, rfpro_software}, limiting accessibility for smaller teams. Additionally, simulation-based testing alone may not suffice without real-world validation, especially in safety-critical domains.

\textcolor{blue}{Deployment readiness varies across platforms and use cases. CARLA dominates academic research, while NVIDIA DRIVE Sim and rFpro are common in industry certification workflows. However, automotive development employs diverse simulation strategies across XiL test benches (Software-in-the-Loop, Model-in-the-Loop, Hardware-in-the-Loop), with tool selection depending on development phase, validation requirements, and organizational infrastructure. Alternative platforms including Gazebo, IPG CarMaker, VTD, and MORAI serve specialized needs across perception development, vehicle dynamics validation, and sensor modeling. This layered approach ensures efficiency, robustness, and regulatory confidence.}

\begin{table*}[htbp]
\centering
\caption{Impact of AV Technologies on Urban Mobility and Infrastructure}
\label{tab:AV_impact_urban_mobility}
\begin{tabularx}{\textwidth}{|X|X|X|X|}
\hline
\textbf{City/Case Study} & \textbf{AV Technology Used} & \textbf{Observed Impacts} & \textbf{Challenges Addressed} \\ \hline
San Francisco~\cite{while2021urban} & Shared AVs & Reduced vehicle ownership, altered parking needs, improved traffic flow & Congestion, urban sprawl, pollution \\ \hline
Singapore~\cite{tan2019adaptive} & Integrated AV and Public Transportation & Enhanced first-mile service, improved public transport efficiency & Congestion, high public transport demand \\ \hline
Toronto~\cite{toronto2022west, bahrami2022autonomous} & AV Ridehailing Operations & Legal adaptation to new transport modes, better city planning & Regulatory frameworks, urban design \\ \hline
Phoenix, Arizona~\cite{mccarroll2022no} & Waymo One Robo-Taxi Service & Increased accessibility, large-scale real-world testing, user acceptance & Safety concerns, insurance policies, liability issues \\ \hline
Pittsburgh, Pennsylvania~\cite{chmielewski2018self} & Autonomous Shuttles and Ridesharing Pilots & Expanded coverage in underserved areas, boosted local robotics ecosystem & Public acceptance, legal complexities, inclement weather adaptations \\ \hline
Hamburg, Germany~\cite{connectedautomateddriving2020autonomous} & Autonomous Shuttle Trials (e.g., HEAT) & Improved integration with public transport, data-driven traffic management & Infrastructure readiness, cybersecurity, operational costs \\ \hline
Dubai, UAE~\cite{while2021urban} & Self-Driving Taxi Trials & Pathway to achieving 25\% driverless trips by 2030, improved road safety & Regulatory approvals, cultural acceptance, extreme climate conditions \\ \hline
Helsinki, Finland~\cite{itf2017shared} & Automated Bus Pilot Projects & Eco-friendly transit solutions, enhanced suburban connectivity & Harsh winter weather, universal design for accessibility \\ \hline
Milton Keynes, UK~\cite{uktin2023mk5g} & StreetCAV Driverless Shuttles & Enhanced urban mobility, reduced congestion, increased public awareness & Public safety, infrastructure integration, regulatory compliance \\ \hline
Sydney, Australia\cite{dowling2022autonomous} & Cit-e Autonomous Vehicle Trials & Improved traffic light coordination, enhanced pedestrian safety & Integration with existing traffic systems, public acceptance \\ \hline
Columbus, Ohio~\cite{smartcolumbus2021sccj} & Smart Columbus AV Shuttle Program & Increased adoption of electric vehicles, improved last-mile connectivity & Public engagement, technological infrastructure, safety validation \\ \hline
Tokyo, Japan~\cite{while2021urban} & Tier IV Self-Driving EV Taxis & Development of spacious autonomous taxis, addressing public transport challenges & Vehicle design, regulatory approval, public acceptance \\ \hline
Masdar City, UAE~\cite{randeree2019social} & Personal Rapid Transit (PRT) System & Demonstrated feasibility of autonomous electric pods for urban transit & Integration with pedestrian traffic, infrastructure costs \\ \hline
Greenwich, London, UK~\cite{TRL_D3.2.8} & GATEway Autonomous Shuttle Trials & Assessed public acceptance and operational challenges of AVs in urban settings & Mixed-use pathways, safety regulations, public perception \\ \hline
Gothenburg, Sweden~\cite{rebalski2024brace} & Volvo's Autonomous Bus Trials & Evaluated performance of large autonomous buses in real traffic conditions & Traffic integration, safety standards, scalability \\ \hline
\end{tabularx}
\end{table*}
\section{Real-World Trials}
\label{sec:autonomus_driving}
Real-world trials represent the most decisive test for autonomous vehicle (AV) systems, exposing limitations that simulations and controlled environments may fail to reveal. Pilot programs across North America, Europe, and Asia are actively assessing how AV technologies, including sensor fusion, machine learning, and connectivity, can contribute to safer roads, reduced emissions, and improved urban mobility~\cite{uk_autodrive,gateway_project,venturer_project,l3pilot_project}. These deployments are expanding the operational scope of AVs in both structured and complex environments, while simultaneously uncovering persistent challenges in perception, decision-making, and safety~\cite{rand_av_policy}. Despite advances in sensing and planning, AVs often struggle with ambiguous layouts, unpredictable road users, and rare events. These shortcomings raise critical questions about liability, certification, and ethical behavior in edge cases. In response, governments are developing legal frameworks and validation procedures, although consensus on standards remains incomplete. Trials have highlighted technical bottlenecks, such as inconsistent lane markings, sensor limitations in adverse conditions, and the need for robust redundancy~\cite{waymo_safety_report,uber_atg}. Regulatory agencies now require transparent disengagement reporting~\cite{rand_av_policy}, while municipalities like Phoenix are focusing on real-world performance in high-risk maneuvers~\cite{mccarroll2022no}. Although trials demonstrate the benefits of AVs in select contexts, such as reduced collisions and enhanced accessibility, they also underscore the complexity of scaling these systems across diverse environments~\cite{zoox_safety,cruise_safety,uk_autodrive,venturer_project,rand_av_policy,nhtsa_ads}. The following sections examine core issues highlighted through these trials, including V2X integration, public trust, evolving regulations, and cross-regional lessons.

\subsection{V2I and V2X Connectivity in Trials}
To achieve safe autonomy in complex settings, AVs must interact not only with their immediate environment but also with external entities such as infrastructure and other vehicles. This is enabled through Vehicle-to-Infrastructure (V2I) and Vehicle-to-Everything (V2X) communication. Real-world trials increasingly leverage these technologies to overcome perception limitations in occluded or low-visibility conditions. For example, roadside units (RSUs) can broadcast signal phase, road hazard alerts, or pedestrian locations beyond an AV’s line of sight. Advances in telecommunications, including 5G and Driving Safety Crash Reporting(DSRC), support low-latency, near-real-time data exchange critical for these applications~\cite{3gpp_cv2x, usdot_v2i, dsrc_cv2x}. Early deployments have demonstrated that V2X connectivity can enhance situational awareness, supplement sensor fusion, and enable safer navigation under challenging conditions. However, deploying V2X systems in real-world settings presents notable challenges. Communication protocols must support both low latency and adequate bandwidth. Trials by MIT’s AV lab and others~\cite{3gpp_cv2x,usdot_v2i} demonstrate that basic messages can be exchanged quickly, but transmitting high-bandwidth data such as raw video remains impractical. As a result, it is more effective to share processed outputs like object locations rather than raw sensor streams. The added data from V2X also increases computational load, requiring AVs to process external messages alongside onboard sensor data within strict timing constraints. Studies~\cite{cv2x_china,dsrc_cv2x} indicate that dedicated V2X modules and efficient GPUs are essential for real-time performance. Several pilot programs are also exploring edge computing, where roadside or cloud units aggregate data from multiple sources and provide summarized information to vehicles. This approach helps reduce the processing burden on individual AVs and allows detection of complex hazards that a single vehicle might miss.

A consistent observation across pilot deployments is that cooperative perception, where vehicles share their sensor data, significantly enhances situational awareness. \textcolor{blue}{OpenCDA~\cite{xu2021opencda} represents a pioneering framework in this domain, providing an integrated co-simulation platform for cooperative driving automation (CDA) research. Building on this foundation, OpenCDA-ROS~\cite{zheng2023opencdaros} enables seamless integration between simulation and real-world cooperative driving automation, facilitating direct deployment of algorithms developed in simulation to physical vehicles through standardized ROS interfaces. CDA-SimBoost~\cite{cdasimboost2024} provides a unified framework specifically designed to bridge real sensor data with simulation environments for infrastructure-based CDA systems, addressing the unique challenges of roadside perception and vehicle-infrastructure coordination.} Systems like COOPERNAUT~\cite{cui2022coopernaut} demonstrate that collaborative communication helps vehicles manage occlusions and long-range detection more effectively than isolated perception. AutoCastSim evaluations showed a 40\% improvement in scenario success rates with only one-fifth the bandwidth of naïve data-sharing methods. COOPERNAUT achieves this by encoding LiDAR data into compact, loss-tolerant messages. Despite promising results, real-world deployment still faces challenges such as unreliable communication, latency variability, and inconsistent standards. Technologies like IEEE 802.11p (DSRC) and Cellular-V2X are being tested globally~\cite{dsrc_cv2x,smart_tl_pilot}. Ensuring data authenticity is also critical, as AVs must trust external information to avoid security risks. Active pilots in Europe and Asia have shown early success. Some cities provide AVs with traffic signal information to optimize timing and reduce stop frequency~\cite{dsrc_cv2x, cress2023intelligent}. In smart city districts, 5G-enabled roadside units enable high-speed platooning and coordinated lane changes~\cite{edge_computing_v2x}. These trials highlight the benefits of V2X is in improving safety through shared awareness and smoother traffic via cooperative behavior. However, scaling requires investment in infrastructure such as smart traffic lights and edge computing nodes, along with standardization across manufacturers. In Phoenix and other testbeds, AVs connected to infrastructure demonstrated reduced idle time and improved reliability~\cite{cooperative_its_corridor}. V2V communication also showed potential to prevent accidents by transmitting early warnings between vehicles. To address bandwidth constraints, current models emphasize efficient data exchange. For example, V2VNet~\cite{wang2020v2vnet} and V2X-ViT~\cite{xu2022v2x} use compact feature aggregation and asynchronous Transformer architectures for cooperative scene understanding. CoBEVT~\cite{xu2022cobevt} enables collaborative bird’s-eye view mapping using sparse fusion techniques and has demonstrated strong performance on the OPV2V dataset. Platforms such as OpenCDA-Loop~\cite{xu2021opencda} enable benchmarking of these models under realistic network conditions, including latency and packet loss. STAMP~\cite{gao2025stamp} offers a task-agnostic framework that integrates multi-agent inputs regardless of model type, facilitating interoperability. Overall, V2X connectivity and cooperative perception are maturing through global trials. While technical and infrastructure challenges remain, early evidence supports their value in improving AV safety and efficiency. Standardization efforts by organizations such as SAE and ISO are likely to play a key role in scaling these systems for broader deployment.
\textcolor{blue}{Beyond pilot programs and cooperative systems, commercial autonomous deployments have reached operational scale. Tesla's Full Self-Driving (Supervised) operates across millions of consumer vehicles, generating large-scale data on edge-case handling and human oversight requirements. Waymo One achieved approximately 4 million fully autonomous paid rides in 2024. This is a ~$\sim$7$\times$ increase over 2023, demonstrating commercial viability within constrained operational design domains.}

\subsection{Trustworthiness, Regulation, and Societal Impact}
\label{subsec:ethical_legal_social}
The successful deployment of autonomous vehicles (AVs) hinges not only on technical maturity but also on addressing critical ethical, legal, and societal challenges. Real-world trials consistently reveal gaps in current regulatory frameworks, emphasizing the need for comprehensive standards that ensure safety, cybersecurity, accountability, and public trust. Foundational safety standards such as ISO 26262~\cite{ISO26262-1:2018} and ISO 21448 (SOTIF)\cite{ISO21448:2022} guide the design of functionally safe systems even under non-failure conditions, while ISO/SAE 21434\cite{ISO-SAE21434:2021} introduces a cybersecurity engineering lifecycle that addresses risks associated with connected vehicle systems. ANSI/UL 4600~\cite{UL4600:2023} complements these by focusing on safety assurance in fully autonomous, driverless products. Recent work such as AutoTrust~\cite{xing2024autotrust} further underscores the importance of trustworthiness benchmarks tailored to AV systems that increasingly incorporate vision-language models (VLMs). These AI-driven architectures, which blend perception with natural language reasoning, introduce novel risks not adequately captured by conventional metrics like disengagement counts or crash frequency. AutoTrust proposes a holistic benchmarking framework that evaluates robustness to adversarial inputs, fairness across demographic and situational contexts, explainability of decision-making processes, and resilience to cyber-physical threats. As AVs become more intelligent and autonomous, these dimensions are critical for validating real-world performance and earning public confidence. Simulation and testing standards are also evolving to support these requirements. ASAM's OpenSCENARIO~\cite{ASAM-OpenSCENARIO:1.3.1,ASAM-OpenSCENARIO:2.0.0}, OpenDRIVE~\cite{ASAM-OpenDRIVE:1.8.1}, OpenODD~\cite{ASAM-OpenODD:Concept}, and OpenLABEL~\cite{ASAM-OpenLABEL:1.0.0} promote scenario interoperability, consistent road modeling, ODD specification, and sensor annotation. As summarized in Table~\ref{tab:av_standards}, these frameworks support a layered governance structure encompassing safety, simulation, and data standards. However, performance evaluation remains fragmented. Metrics such as those reported to the California DMV or under NHTSA's Standing General Order~\cite{NHTSA-SGO:2021-01} lack global consistency and fail to capture higher-order trust dimensions. Frameworks like RAVE~\cite{Scanlon2024RAVE} enable human-AV performance comparison, yet the absence of a unified "autonomy score" limits cross-system benchmarking. International efforts by UNECE including UN-R157~\cite{UN-R157:2021} for Level 3 automation, UN-R155~\cite{UN-R155:2021} for cybersecurity, and UN-R156~\cite{UN-R156:2021} for software updates represent critical steps toward harmonized regulation but require further development to support higher automation levels and diverse urban deployments.
Legal and governance frameworks must advance alongside technical standards to ensure responsible AV deployment. Most existing traffic laws assume a human driver, which creates ambiguity when control shifts to automated systems. Jurisdictions such as Arizona and Toronto have updated their definitions to classify the automated driving system (ADS) as the legal driver and now require companies to provide financial guarantees through insurance or bonds~\cite{mccarroll2022no,uktin2023mk5g}. These policy adaptations reflect an incremental learning process based on real-world trials, although global agreement on liability structures is still evolving~\cite{chmielewski2018self}. As AVs move toward commercial deployment, responsibility may shift toward manufacturers under product liability frameworks, which will require strong documentation, audit trails, and compliance mechanisms. Ethical dilemmas have also surfaced in real-world deployments. The 2018 Uber ATG crash in Tempe, Arizona, illustrated how AV design decisions can encode implicit value judgments, especially in unavoidable harm scenarios~\cite{mccarroll2022no}. In response, cities like Hamburg and London have incorporated public input into policy formation. Hamburg's HEAT shuttle project and London's GATEway initiative involved citizen panels and community workshops to guide AV behavior and expectations~\cite{connectedautomateddriving2020autonomous,TRL_D3.2.8}. Transparent reporting is vital to maintaining trust. Studies in Pittsburgh and Tempe found that lack of transparency can reduce public support by up to 80 percent, while active community engagement in Phoenix and Columbus has improved acceptance~\cite{chmielewski2018self,smartcolumbus2021sccj}.

\textcolor{blue}
{AV deployment requires integrated management across four dimensions. \textit{Liability:} Transition from driver-based tort law to manufacturer product liability necessitates clear documentation chains and OTA update accountability such as the Arizona's ADS-as-driver classification~\cite{mccarroll2022no} exemplifies evolving frameworks. \textit{Transparency:} Black-box AI decision-making undermines trust; explainability requirements (AutoTrust~\cite{xing2024autotrust}) and public disengagement reporting (California DMV, NHTSA SGO~\cite{NHTSA-SGO:2021-01}) establish baseline disclosure standards. \textit{Fairness:} Algorithmic bias risks discriminatory outcomes. For example: Pittsburgh and Columbus trials emphasize equitable service distribution across socioeconomic zones~\cite{chmielewski2018self,smartcolumbus2021sccj}. \textit{Public Trust:} Hamburg and London projects demonstrate that citizen engagement during development increases acceptance by 60-80\% compared to post-deployment transparency alone~\cite{connectedautomateddriving2020autonomous,TRL_D3.2.8}. These dimensions must evolve in parallel with technical standards to ensure responsible scaling.}
Wider societal impacts also warrant attention. Trials in Columbus and Pittsburgh have shown that AVs can improve mobility for underserved populations~\cite{chmielewski2018self,smartcolumbus2021sccj}. However, concerns persist around job displacement and unequal service distribution. In San Francisco, critics noted that early deployments favored high-income areas~\cite{while2021urban}, prompting discussions on equitable rollout and worker transition programs. Public reaction remains mixed. Sydney's AV trials received widespread interest and support~\cite{dowling2022autonomous}, while opposition in San Francisco has included protests and obstruction of vehicles. These findings highlight the importance of transparency, fairness, and public engagement as integral components of successful AV integration.

\subsection{Pathways to Overcoming Current Limitations and Exploiting New Opportunities} \label{subsec:pathways} 
Table~\ref{tab:AV_impact_urban_mobility} summarizes global case studies of AV deployment, highlighting diverse challenges and strategies. Several clear pathways have emerged. Infrastructure investment is a major factor: cities such as Singapore and Dubai, which have implemented advanced V2I systems, report 90 to 95 percent technical success and over 80 percent public acceptance~\cite{tan2019adaptive, while2021urban}. In contrast, cities with limited infrastructure upgrades, including Phoenix and Pittsburgh, report 85 to 90 percent success but experience longer deployment timelines and increased regulatory uncertainty~\cite{mccarroll2022no, chmielewski2018self}. However, data suggests that the return on investment diminishes beyond 20 million USD per deployment zone. This points to the efficiency of targeted upgrades like smart intersections and V2X systems over fully integrated infrastructure. Urban trials have benefited from integrating AVs with public transport and adaptive signal systems, as seen in Singapore~\cite{tan2019adaptive}. Regulatory strategies in Dubai and Hamburg have also emphasized cybersecurity readiness and liability definitions~\cite{while2021urban, connectedautomateddriving2020autonomous}. Performance analysis shows that constrained, geofenced environments deliver up to 98 percent success within design limits but perform significantly worse outside those zones. Dense urban deployments maintain 85 to 90 percent reliability under normal traffic but fall to 60 to 70 percent during peak hours or construction~\cite{itf2017shared}. New business models are also shaping AV rollouts. Robo-taxi services in Phoenix~\cite{mccarroll2022no} and ride-hailing initiatives in Toronto~\cite{toronto2022west, bahrami2022autonomous} illustrate shifts toward subscription and pay-per-use models. Suburban services can reach break-even with under 10,000 rides per vehicle per month, while urban operations require significantly higher utilization due to cost and complexity. Waymo's 2024 record of 4 million paid trips, a sevenfold increase from 2023, shows commercial momentum. Conversely, GM’s exit from Cruise illustrates the financial burden of AV deployment, with validation costs often exceeding 1 to 5 million USD per automaker. Public trust is a crucial enabler. Hands-on AV experiences in cities like Pittsburgh~\cite{chmielewski2018self} and Greenwich~\cite{TRL_D3.2.8} have enhanced acceptance and enabled productive feedback loops between communities, planners, and developers. Proactive engagement with regulators can reduce deployment delays by up to 80 percent. Public education initiatives launched within 12 months of deployment have raised acceptance by approximately 70 percent. Transparent incident reporting helps maintain trust, while lack of disclosure can reduce support by as much as 80 percent. In conclusion, successful AV deployment depends on targeted infrastructure investment, adaptive regulation, inclusive economic strategies, and consistent community engagement. The intersection of mature technology, formalized validation frameworks, and sustained public support marks the path toward safe and scalable AV integration.

\begin{table*}[htbp]
\centering
\caption{Current Standards and Metrics for Autonomous Vehicle Development and Simulation}
\label{tab:av_standards}
\footnotesize
\begin{tabular}{|p{3cm}|p{1.5cm}|p{3cm}|p{4cm}|p{3.5cm}|}
\hline
\textbf{Standard/Metric} & \textbf{Organization} & \textbf{Domain} & \textbf{Purpose \& Scope} & \textbf{Key Applications} \\
\hline
\multicolumn{5}{|c|}{\textbf{Safety and Functional Standards}} \\
\hline
ISO 26262:2018\cite{ISO26262-1:2018} & ISO TC22/SC32 & Functional Safety & Road vehicle functional safety for E/E systems. Addresses systematic and random hardware failures through ASIL (A-D) risk classification & ADAS development, ECU validation, safety-critical system design \\
\hline
ISO 21448:2022 (SOTIF)\cite{ISO21448:2022} & ISO TC22/SC32 & Safety of Intended Functionality & Addresses hazards from functional insufficiencies and foreseeable misuse without system failures. Complements ISO 26262 & AI/ML validation, autonomous driving systems, edge case scenarios \\
\hline
ISO/SAE 21434:2021\cite{ISO-SAE21434:2021} & ISO/SAE & Cybersecurity & Automotive cybersecurity engineering lifecycle. Risk assessment (TARA), security goals, and continuous monitoring & Connected vehicles, OTA updates, cyber threat mitigation \\
\hline
ANSI/UL 4600:2023\cite{UL4600:2023} & UL & Autonomous Product Safety & Safety evaluation for autonomous products beyond traditional functional safety & Autonomous vehicles, robotics, safety certification \\
\hline
\multicolumn{5}{|c|}{\textbf{Automation Level Classification}} \\
\hline
SAE J3016:2021\cite{SAE-J3016:2021} & SAE/ISO & Driving Automation Levels & Taxonomy for 6 levels (0-5) of driving automation. Defines DDT, ODD, fallback responsibilities & AV classification, regulatory frameworks, ODD definition \\
\hline
\multicolumn{5}{|c|}{\textbf{Simulation and Testing Standards}} \\
\hline
ASAM OpenSCENARIO 1.3.1\cite{ASAM-OpenSCENARIO:1.3.1} & ASAM & Dynamic Simulation Content & XML-based format for describing dynamic driving scenarios with multiple entities and complex maneuvers & Scenario-based testing, ADAS validation, simulation interoperability \\
\hline
ASAM OpenSCENARIO 2.0\cite{ASAM-OpenSCENARIO:2.0.0} & ASAM & Domain-Specific Language & Declarative language for abstract scenario descriptions with parameterization and coverage specification & Test generation, stochastic scenarios, automated validation \\
\hline
ASAM OpenDRIVE 1.8.1\cite{ASAM-OpenDRIVE:1.8.1} & ASAM & Road Network Description & XML format for static road network descriptions including geometry, lanes, signals, and junctions & HD map creation, simulation environments, road network exchange \\
\hline
ASAM OpenCRG 1.1.2\cite{OpenCRG:1.1.2} & ASAM & Road Surface Description & Curved regular grid format for high-precision road surface elevation and properties data & Tire simulation, vehicle dynamics, surface modeling \\
\hline
ASAM OSI 3.x\cite{ASAM-OSI:3.7.0} & ASAM & Simulation Interface & Standardized interfaces between simulation models and components for distributed simulations & Sensor simulation, environment perception, tool integration \\
\hline
\multicolumn{5}{|c|}{\textbf{Performance and Evaluation Metrics}} \\
\hline
NHTSA Standing General Order\cite{NHTSA-SGO:2021-01} & NHTSA & Crash Reporting & Mandatory reporting of AV crashes involving injury, fatality, or significant property damage & Safety monitoring, performance benchmarking, regulatory compliance \\
\hline
RAVE Framework\cite{Scanlon2024RAVE} & Industry Consortium & Retrospective AV Evaluation & Best practices for comparing AV and human driving performance using crash data analysis & Safety impact assessment, benchmarking methodologies \\
\hline
\multicolumn{5}{|c|}{\textbf{Emerging and Specialized Standards}} \\
\hline
ISO 34502:2022\cite{ISO34502:2022} & ISO & Road Vehicle Data & Data specifications for intelligent transport systems and automated driving functions & Data exchange, interoperability, system integration \\
\hline
IEEE 2846:2022\cite{IEEE2846:2022} & IEEE & Assumptions for AV Safety & Assumptions for mathematical models used in safety-related automated vehicle behavior & Model validation, safety argumentation, formal verification \\
\hline
ASAM OpenLABEL 1.0\cite{ASAM-OpenLABEL:1.0.0} & ASAM & Sensor Data Annotation & Standardized format for annotating sensor data (camera, LiDAR, radar) for ML training & Dataset annotation, ML model training, validation \\
\hline
ASAM OpenODD\cite{ASAM-OpenODD:Concept} & ASAM & Operational Design Domain & Standardized description of operational design domains for automated driving systems & ODD specification, system limitations, safety boundaries \\
\hline
\multicolumn{5}{|c|}{\textbf{International Harmonization}} \\
\hline
UN-R157 ALKS\cite{UN-R157:2021} & UNECE WP.29 & Automated Lane Keeping & International regulation for automated lane keeping systems on highways (Level 3) & Highway automation, international deployment, type approval \\
\hline
UN-R155 Cybersecurity\cite{UN-R155:2021} & UNECE WP.29 & Vehicle Cybersecurity & Cybersecurity and software update management systems for road vehicles & Type approval, cybersecurity management, international standards \\
\hline
UN-R156 Software Updates\cite{UN-R156:2021} & UNECE WP.29 & Software Update Management & Over-the-air software update systems and management for road vehicles & OTA updates, software lifecycle, regulatory approval \\
\hline
\end{tabular}
\end{table*}

\section{Discussion: Envisioning the Future of Autonomous Driving}
\label{discussion}
\textcolor{blue}{While our survey has focused on simulation-to-real-world deployment, the following forward-looking approaches represent architectural and methodological innovations essential for bridging gaps identified throughout this work particularly in Sim2Real transfer, digital twin validation, and real-world trial scalability.} AVs must evolve beyond visual perception to reasoning about semantics, causality, and uncertainty~\cite{sadigh2016planning,grigorescu2020survey}. Hybrid approaches such as neuro-symbolic systems offer promise by combining deep learning with logical reasoning~\cite{de2011neural,landajuela2021discovering}. System-level integration is also shifting toward decentralized cooperation. Trials of V2V and V2X communication have shown up to 40\% gains in occluded and long-range perception~\cite{cui2022coopernaut,xu2022cobevt}, but reliable communication and consensus protocols remain essential for scale. Digital twins are central to continuous AV validation, especially when synchronized with real-world data~\cite{wang2024digital,digitaltwinconcepts2022}. Emerging approaches now embed human behavior into simulations to improve negotiation and planning in complex traffic~\cite{wei2024editable}. Despite progress in standards like ISO 26262 and UNECE regulations, deployment is hindered by gaps in V2X consistency and safety metrics~\cite{dosovitskiy2017carla,sarker2024comprehensive,3gpp_cv2x}. Traditional metrics like disengagements do not fully capture system risk, motivating more scenario-driven and adversarial testing frameworks~\cite{revell2022sim2real,xu2022v2x}. Finally, the sim-to-real gap remains a key obstacle. While simulations help generate edge cases, current models struggle to transfer reliably without domain adaptation~\cite{zhao2023real2sim2real,sim2realcare,yu2024natural}. Bridging this gap will be vital for safe and scalable AV deployment.

Several forward-looking approaches show strong potential to overcome current AV limitations. One is the integration of neuro-symbolic AV stacks, which combine neural networks for perception and control with symbolic reasoning for planning and scenario understanding. This hybrid method enables AVs to generalize from data while reasoning about constraints and intent, improving interpretability and regulatory transparency~\cite{eom2024selective}. For instance, such a system could recognize an emergency vehicle using neural vision and apply logical rules to yield, even in novel situations~\cite{de2011neural,landajuela2021discovering}. A second direction involves federated fleet learning. Unlike traditional cloud-based updates, federated learning allows vehicles to share model parameters without transmitting raw data. This facilitates real-time adaptation to local traffic conditions, such as roadwork or unusual pedestrian behavior, while preserving privacy and minimizing bandwidth. Fleet-wide learning from rare edge cases becomes feasible, supporting collective resilience. The third area is self-adaptive digital twins. These systems continuously synchronize with live AV data to update their simulation state, road layout, and agent behaviors. Such adaptive twins support scenario replay, predictive diagnostics, and closed-loop training. For example, anomalous pedestrian movements encountered in real driving can be re-simulated with variations to improve avoidance strategies~\cite{wang2024digital,digitaltwinconcepts2022}. Finally, zero-shot Sim2Real transfer aims to enable models trained solely in simulation to perform reliably in real-world deployments without fine-tuning. This requires high domain diversity through procedural content generation and realism in sensor and physics modeling. Long-term goals include using foundation models trained across multiple simulation domains to support robust, generalizable AV behavior~\cite{yu2024natural,salvato2022sim}. To advance safe and scalable AV deployment, several actionable insights and open challenges should guide future research:
\begin{itemize}
    \item \textbf{Hybrid Reasoning Architectures:} AVs should integrate neural learning with symbolic logic to improve interpretability, safety, and compliance. Logic-based rules can constrain unsafe behavior and enable formal verification, critical for regulated deployment~\cite{sadigh2016planning,landajuela2021discovering}.
    \item \textbf{Human-in-the-Loop Simulation:} Embedding realistic human behaviors into digital twins allows testing of nuanced interactions such as eye contact or unexpected pedestrian actions. This supports safer policy learning and interactive debugging without physical risk~\cite{wei2024editable,lee2024ad4rl}.
    \item \textbf{Cooperative Autonomy:} To handle occlusions and dynamic traffic, AVs must share perception, intent, and planning cues through V2X. Future work should address robust multi-agent fusion, low-latency communication, and fail-safe coordination~\cite{cui2022coopernaut,xu2022cobevt}.
    \item \textbf{Regulatory Synchronization:} Fragmented standards in safety, connectivity, and liability pose barriers to deployment. Harmonizing protocols, establishing performance metrics, and integrating runtime traceability into AV stacks are urgent priorities~\cite{dosovitskiy2017carla,3gpp_cv2x}.
    \item \textbf{Zero-Shot Sim2Real Transfer:} Generalizing to unseen real-world domains without fine-tuning remains a major bottleneck. Progress requires better domain-invariant representations, curriculum design, and rigorous benchmarking across diverse conditions~\cite{yu2024natural,salvato2022sim}.
\end{itemize}
These challenges reflect both the technical complexity and societal integration required for large-scale AV adoption. Sustained collaboration between researchers, industry, and policymakers will be essential to navigate these next stages effectively.

\section{Conclusion}
\label{conclusion}
This survey has explored the growing role of synthetic data and simulation technologies in advancing autonomous driving. With real-world testing often limited by cost and safety constraints, virtual environments, procedural content generation, and digital twins offer scalable alternatives for developing and validating AV systems. We reviewed how synthetic datasets support perception and planning tasks, and how digital twins enable system-level evaluation through closed-loop and human-in-the-loop testing. The integration of domain adaptation techniques and foundation models, including vision-language frameworks, points to increasingly transferable and robust AV capabilities. Key challenges persist, particularly in achieving reliable Sim2Real transfer, establishing standardized validation protocols, and ensuring safety in rare or ambiguous scenarios. Addressing these issues will require continued progress in simulation fidelity, multi-agent coordination, and learning efficiency. Emerging research directions such as adaptive digital twins, decentralized cooperative autonomy, and neuro-symbolic decision-making offer promising pathways forward. By closing the gap between synthetic and real-world domains, the community can accelerate the deployment of safe and scalable autonomous driving systems.

\section*{Acknowledgments}
\noindent This research is supported by the UK Research and Innovation (UKRI)—Engineering and Physical Sciences Research Council (EPSRC) Grant ATRACT (EP/X028631/1).
\bibliographystyle{IEEEtran}
\bibliography{mybibliography} 
 \vspace{-60pt}
\begin{IEEEbiography}[{\includegraphics[width=1in, height=1.25in, clip, keepaspectratio]{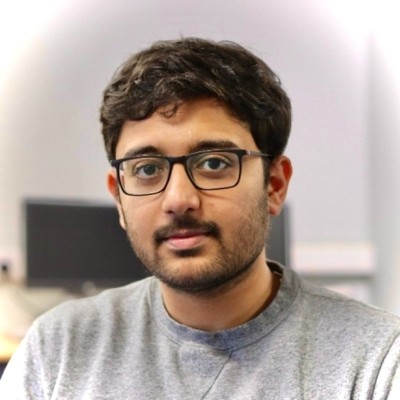}}]{Aditya Humnabadkar}
(Student Member, IEEE) received his M.Sc. in Smart, Connected and Autonomous Vehicles from the University of Warwick, UK, in 2021, and completing his Ph.D. in Computer Science at Edge Hill University, UK. He is a Data Scientist at the Office for National Statistics, UK. His research interests include deep learning, computer vision, and multimodal AI, with applications in automotive, economics, and finance. \vspace{-60pt} \end{IEEEbiography}

\begin{IEEEbiography}[{\includegraphics[width=1in, height=1.25in, clip, keepaspectratio]{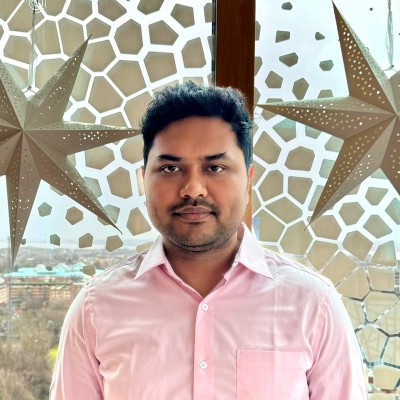}}]{Arindam Sikdar} (Member, IEEE) received his M.Tech in Electronic and Communication Engineering from IIT Bhubaneswar, India, in 2015, and his Ph.D. from Jadavpur University, Kolkata, in 2021. He worked as a postdoctoral fellow at the University of Bristol, UK, contributing to the SPHERE project, and is currently a postdoctoral researcher at Edge Hill University, UK. His research interests include computer vision, pattern recognition, visual surveillance, multimedia analysis, and digital healthcare. \vspace{-60pt}
\end{IEEEbiography}

\begin{IEEEbiography}[{\includegraphics[width=1in, height=1.25in, clip, keepaspectratio]{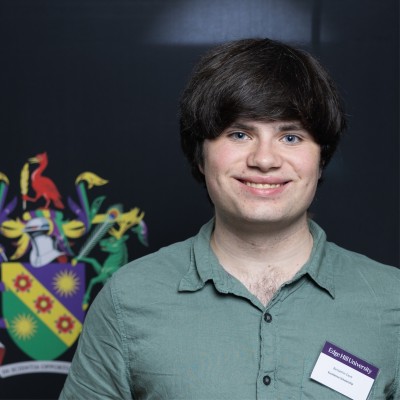}}]{Benjamin Cave} is a final-year undergraduate student in Computer Science and Artificial Intelligence at Edge Hill University, UK. His academic interests span machine learning, data analytics, and visual recognition, with a specialisation in video-based feature extraction. His current research focuses on the development of intelligent systems for visual recognition tasks. \vspace{-60pt} \end{IEEEbiography}

\begin{IEEEbiography}[{\includegraphics[width=1in, height=1.25in, clip, keepaspectratio]{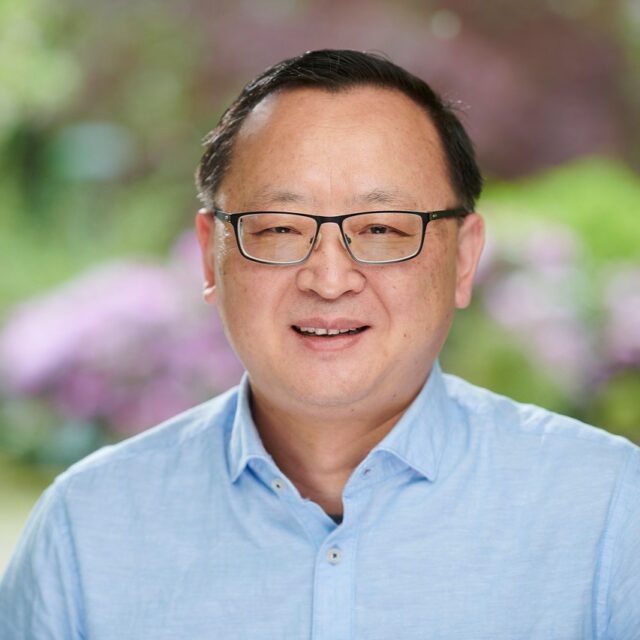}}]{Huaizhong Zhang} received his B.Sc. from East China Normal University, China, his M.Sc. from Southeast University, China, and his Ph.D. from the University of Ulster, UK, in 2010. He is currently a Senior Lecturer in the Department of Computer Science at Edge Hill University, UK. His research interests include deep learning, big data, medical image analysis, pattern recognition, and object tracking and recognition. \vspace{-60pt} \end{IEEEbiography}

\begin{IEEEbiography}[{\includegraphics[width=1in, height=1.25in, clip, keepaspectratio]{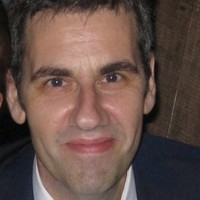}}]{Nik Bessis} 
(Senior Member, IEEE) received his BA from the TEI Athens (1991) and his MA and PhD degrees from De Montfort University, UK in 1995 and 2002, respectively. He is a full Professor of Computer Science (2010-). He established and since 2015, was the Director of Data Science research centre, Head (Chair) of the Department of Computer Science and the Department of Engineering at Edge Hill University, UK. Since 2021, he is a University Senior Advisor (Research). In 2024, he established the Data Science \& AI multi-institutional research alliance for global \& societal challenges. He is an Institutional Lead for UKRN, a member of the EDI advisory group and EUA, CDE. His research is on IoT, smart systems, social graphs for network and big data analytics. He is involved in projects worth over £15m. Prof Bessis has published over 350 works and won 4 best papers awards. \vspace{-60pt}
\end{IEEEbiography}

\begin{IEEEbiography}[{\includegraphics[width=1in, height=1.25in, clip, keepaspectratio]{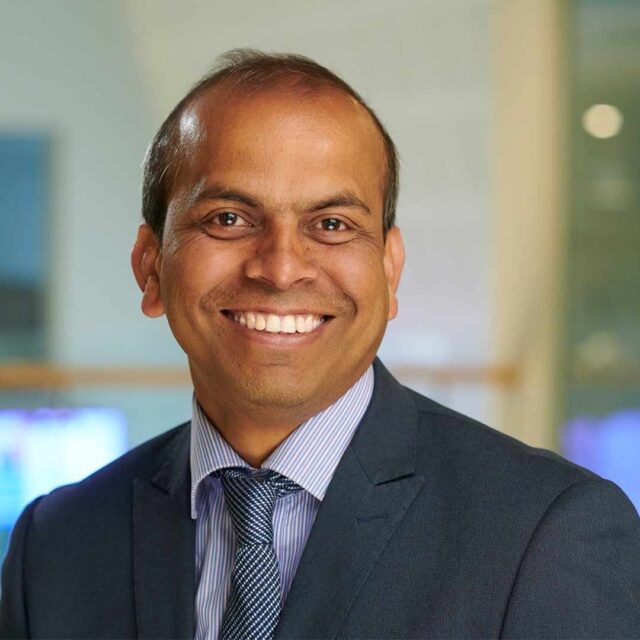}}]{Ardhendu Behera} (Member, IEEE) is a Professor of Computer Vision and AI and Director of the Intelligent Visual Computing Research Centre at Edge Hill University, UK \& Deputy Director for HRI (Health Research Institute, UK). He serves as an Associate Editor for IEEE Transactions on Image Processing. His research interests include deep learning, computer vision, and multimodal AI, with applications in automotive, defence, and healthcare.
\end{IEEEbiography}
\end{document}